\documentclass[applsci,article,accept,moreauthors,pdftex]{Definitions/mdpi} 
\usepackage[]{algorithm2e}
\usepackage{amsmath}
\usepackage{tabularx}
\usepackage{multicol}
\usepackage{comment}
\usepackage{array}
\usepackage{mathtools}
\usepackage{upgreek} 
 \usepackage{textcomp}

\firstpage{1} 
\pubvolume{xx}
\issuenum{1}
\articlenumber{5}
\pubyear{2020}
\copyrightyear{2020}
\history{Received: date; Accepted: date; Published: date}
\updates{yes} % If there is an~update available, un-comment this line

\Title{An %Attention AE/ME. The following layout issues have not been checked by the English Editing Department and must be carefully verified by the AE/Layout Department: All callout issues, bold usage of callouts, and references to callouts in the text. Correct callout usage in figures. Figure and Table layout issues. Footnote formatting and Glossaries have not been checked. En dash usage for negative values, en dash usage to indicate relationships, en dash usage to indicate bonds (especially in chemistry). The English Editing Department is not responsible for correct italic usage for genes, proteins and technical terminology. This responsibility belongs to the authors. The following are also not checked: spacing between numbers and units of measurement, ratios, en dashes for ranges, date and time formats, punctuation in equation lines, and less than/more than spacing (< >). Finally, capitalization and layout of titles/headings must be properly checked as well as ensuring 'Eq.' and 'Fig.' are properly spelled out, as these are layout issues.
	Intelligent System-on-a-Chip for a~Real-Time Assessment of Fuel Consumption to Promote Eco-Driving}

\Author{Óscar Mata-Carballeira $^{*}$\orcidA{}, Mikel Díaz-Rodríguez, Inés del Campo, and Victoria Martínez}

\AuthorNames{Óscar Mata-Carballeira, Mikel Díaz-Rodríguez, Inés del Campo and Victoria Martínez}

\address[1]{Department of Electricity and Electronics, Faculty of Science and Technology, University of the Basque Country UPV/EHU, 48940 Leioa, Spain; mdiaz063@ikasle.ehu.eus (M.D.-R.); ines.delcampo@ehu.eus (I.d.C.); victoria.martinez@ehu.eus (V.M.)}

\corres{Correspondence: oscar.mata@ehu.eus; Tel.: +34-94-601-3238}

\abstract{%Pollution originated by automobiles is a~concern in the current world, not only because of   global warming, but also due to the harmful effects on people's health and lives. Despite strict regulations on exhaust gas emissions being applied, minimizing unsuitable driving habits, associated with elevated fuel consumption and emissions, would achieve further reductions. For that reason, this work proposes a~self-organized map  (SOM)-based intelligent system to provide drivers with eco-driving-intended driving style (DS) recommendations. The development of the DS advisor uses driving data from the Uyanik instrumented car, which includes data from a~CAN-bus, an~inertial measurement unit, and a~laser ranging unit. The system classifies drivers into five different DS classes regarding how ecological their driving behaviors are. When compared to other implementations, the main advantage of this system is that recommendations are provided to motorists with comprehensible language according to their driving characteristics, contributing to encouraging eco-driving behaviors, with potential improvements of up to the 41\% in both fuel consumption and emissions. The system was successfully implemented using a~field-programmable gate array (FPGA) device of the Xilinx Zynq programmable system-on-a-chip (PSoC). This SOM-based system allows for  high-speed real-time implementation, achieving state-of-the-art timing performances and low power consumption. It is suitable for the development of on-board advanced driving assistance systems (ADASs).}
Pollution that originates ftom automobiles is a~concern in the current world, not only because of global warming, but also due to the harmful effects on people's health and lives. Despite regulations on exhaust gas emissions being applied, minimizing unsuitable driving habits that cause elevated fuel consumption and emissions would achieve further reductions. For that reason, this work proposes a~self-organized map (SOM)-based intelligent system in order to provide drivers with eco-driving-intended driving style (DS) recommendations. The development of the DS advisor uses driving data from the Uyanik instrumented car. The system classifies drivers regarding the underlying causes of non-optimal DSs from the eco-driving viewpoint. When compared with other solutions, the main advantage of this approach is the personalization of the recommendations that are provided to motorists, comprising the handling of the pedals and the gearbox, with potential improvements in both fuel consumption and emissions ranging from the 9.5\% to the 31.5\%, or~even higher for drivers that are strongly engaged with the system. It was successfully implemented using a~field-programmable gate array (FPGA) device of the Xilinx ZynQ programmable system-on-a-chip (PSoC) family. This~SOM-based system allows for real-time implementation, state-of-the-art timing performances, and low power consumption, which are suitable for developing advanced driving assistance systems (ADASs).} %200 palabras


\begin{document}
%%%%%%%%%%%%%%%%%%%%%%%%%%%%%%%%%%%%%%%%%%

%%%%%%%%%%%%%%%%%%%%%%%%%%%%%%%%%%%%%%%%%%

\section{Introduction}

Throughout the years, the~paradigm of transportation has experienced several changes. At~the dawn of the transportation era, vehicles were designed for serviceability and exclusiveness as the main objectives, no matter how polluting the engine was. This perspective gradually changed to make the automobile widespread among the entire population; however, fuel economy was not a~real concern because fossil fuels were cheap and abundant. However, in~the 1970s, the~latter approach of big cars moved by massive engines needed to be put apart due to the rise in fuel prices~\cite{daito2000automation}. It was in this time when car manufacturers set the basements for engine downsizing aiming to the obtainment of acceptable fuel economy rates, but~pollution and global warming were not still a~main concern~\cite{stern1977air}. 

It was in the 1980s when pollution became a~main issue for both the public   and for the developed countries' health systems~\cite{Chappie1982,andersen2020air}. Several pieces of research observed %please check
 that pollution mainly contributed to climate change and worsened the health condition  of people suffering from breathing system pathologies in cities with high emission levels~\cite{Evans1981}. 

For those reasons, alternative energy-based means of transportation, which mainly rely on electric energy, emerged as a~way to mitigate those harmful effects. However, because electricity-powered automobiles needed an~%please check
 evolution in both power storage and power train technologies, hybrid~cars arose as a~trade-off solution between electric and petrol cars. Despite hybrid automobiles still being a~polluting means of transportation, the~addition of electric engines into their power train contributes to increasing their power efficiency and, consequently, their eco-friendliness~\cite{Onat2015}. Nevertheless,~the~outstanding advances in battery technology and energy converters in recent years have changed the paradigm of non-polluting transportation, since they made electric cars a~market reality~\cite{Ahmad2018,Haddadian2015}. Since the early 2010s, electric cars' sales have consistently risen, in~countries, such as Norway, where this kind of automobile is mainly sold~\cite{aarstad2020has}. These cars, though, still have several problems, such as the impact of their complete life cycle. Though~advances in recycling several components, such as batteries~\cite{Hanisch2015}, have been made, the~use of energy-intensive materials causes a~noticeable impact on greenhouse gas emissions~\cite{zehner2013unclean,Haddadian2015}. On~the other hand, at~least for the current decade, fuel-powered cars are expected to continue as the mobility standard~\cite{Ortar2019,Morgan2019}. Thus,~the~development of systems to change low-efficiency driving behaviors, such as driving at high regimes, seems to be a~good alternative in efforts to increase the eco-friendliness of the current fleet of fuel-powered~cars.

\subsection{Pollution and Environmental~Regulations}
Although emissions from transport are not the main air pollution source in all major cities of the world~\cite{Paton-Walsh2019}, the~inhabitants of many cities worldwide suffer the effects of these gasses. Particularly,~emissions from private transportation cause 5\% of the 3--4 million deaths in Europe and the US directly related to general air pollution~\cite{Nieuwenhuijsen2016, Hoffmann2018}. Private transportation plays an~important role in emitting toxic gases identified in urban areas’ air, with~consequences in both citizens' health and global warming~\cite{sun2018urban}. With~the aim of reducing these effects, several local governments have restricted private traffic in urban areas, such as London with its Ultra-Low Emissions Zone (ULEZ) \cite{Ellison2013, Pettit2020} or the Madrid Central LEZ~\cite{Salas2019}. Additionally, national and transnational institutions have deployed environmental regulations that force car manufacturers to develop more ecological automobiles~\cite{mcbain2018reducing, AWAD2020137302}.

Those standards have had a~positive effect, with~CO emissions reduced up to 82\%, HC down 50\%, NOx 84\%, and~PM emitted by diesel down 96\%. Nevertheless, EU authorities have pointed out that NOx emissions have not been reduced as much as expected, particularly in the case of diesel vehicles, since homologation tests provide results that are poorer than real-world measurements~\cite{Hooftman2018}. 

\begin{comment}

\subsubsection{Local~Policies}

To address the aforementioned problem, local authorities have implemented complementary restrictions on traffic in urban areas. It is well known that different cities have come out with several programs so as to cope with the poor air quality and the derived health problems.  These programs vary in the way the restrictions are applied, namely economically or unconditionally. While economical programs rely on charging toll fees to drivers of incoming vehicles, unconditional ones are based only on ecological premises. A~particularly well known economical program is the London's Low Emission Zone (LEZ), aimed to reduce the impact of exhaust gases in the Greater London area by restricting the circulation of vehicles non-compliant with the EURO 4 standard  since  January 2012~\cite{Ellison2013}. Since April 2019, these restrictions have applied  at all times~\cite{Pettit2020}. To~enter this zone, a~toll is applied regarding the class and the eco-friendliness of the vehicle. This measure, since its implementation, has reduced emissions by more than 20\%. On~the other hand, we can find the Madrid Central LEZ, implemented in 2018, as~an example of an~unconditional restricted zone~\cite{Salas2019}. In~this case,   residents can freely circulate downtown Madrid as long as their vehicles are compliant with the regulations. Non-residents, however, are only allowed to enter in the case they drive to a~parking lot. These restrictions are applied with no~exceptions. 

\end{comment}

\subsection{Related~Research}
\label{sec:Related_res}

Even though traffic restrictions and environmentally friendly means of transportation are a~reality, their effects on reducing greenhouse effect gasses have been found to not be as significant as expected~\cite{zehner2013unclean}. In~that sense, it has been observed that individuals' DS plays a~more important role in emitting polluting agents than the ecological rating of the vehicle itself, with~studies showing that, in~different situations, aggressive driving could increase energy consumption by 47\% \cite{JAVANMARDI201713866}. With~these assertions in mind, it seems reasonable that if we could assess the fuel-consumption efficiency of individuals, their DS could be corrected in order to increase their ecological~friendliness.

As found in several studies, the~manner motorists operate the throttle and brake pedals, their desired rate of acceleration, speed control, and~control stability play a~major role in fuel consumption, regardless of the driven vehicle. Thus, we can learn how some drivers have a~higher energy cost than others by studying the impact of their driving behavior on fuel consumption, thus helping high-energy-cost drivers to achieve energy-efficient DS. Factors, such as personality, ability and skills, attitudes, perceptions, socio-economic characteristics, age, gender, and~experience, among~others, have been identified to be related to riskier and more aggressive driving events, such as extreme accelerations, excess revolutions per minute (RPM), extreme braking, and hard starts, events that cause high fuel consumption~\cite{rolim2017real}.

It should be pointed out that eco-driving is mainly an~operational decision that allows drivers to maximize fuel efficiency and reduce pollutants' emissions. It is characterized by the use of several techniques that help to maximize the vehicle's energy efficiency. Therefore, this concept can be seen as a~set of rules that differ from the driving that motorists are used to performing, including calm driving, the~avoidance of unnecessary stops, and~the anticipation and elimination of idling when possible. %Authors have identified key factors influencing eco-driving, such as weather, vehicle, roadway, traffic, and~drivers’ characteristics in relation to fuel consumption~\cite{LOIS2019232}. On~the other hand, two approaches on fuel efficiency-intended driving studies can be identified: heuristic strategies and model-based strategies. The~former are based on eco-driving rules such as   the aggressiveness of   acceleration,  the~stability of the cruising speed,  the~anticipation and aggressiveness of   deceleration and braking,  the~duration of idling, and~optimal gear changing. The~latter use optimal control techniques to minimize a~cost function that depends basically on  physical constraints, system dynamics, and~other parameters, such as speed limits~\cite{ojeda2017real}. 
Several authors remark that eco-driving could effectively contribute to reducing overall fuel consumption and CO$_2$ emissions if adequate education about strategic, tactical, and~operational decisions were provided to drivers~\cite{sivak2012eco, ayyildiz2017reducing, Gilman2018}. In~this sense, during~the trip, and~when the trip has finished, providing practical recommendations might be~useful. 

In the most commonly used form of eco-driving measures, drivers are given advice in training sessions, and~the organizers measure differences in fuel consumption and CO$_2$ emissions before and after training~\cite{Baric2013}. Another valid approach is providing a~report of the strengths and weaknesses after each eco-driving session~\cite{Allison2019}. Nevertheless, a~natural evolution on those lessons provides instantaneous feedback of the driver's operational decisions~\cite{Allison2019}. It has been found that on-trip eco-drive support is more efficient, with~reductions of up to 10\% on fuel consumption when compared to post-drive assessment (which only achieves a~5\% reduction) \cite{ayyildiz2017reducing}. However, the~former is more expensive and it requires complex algorithms as well as real-time technology dependence, while the latter can be provided through an~end-of-trip fuel consumption assessment~\cite{ayyildiz2017reducing}. Thus, the~impact of eco-drive education has been consistently verified in the reference publications, from~particular drivers to transportation~professionals.

For that reason, several attempts of fuel economy-intended system implementations, acting~on the aforementioned parameters, such as gear recommendation~\cite{Long2019} or eco-driving scoring, have~been deployed in cars~\cite{Tulusan2012}. These systems, despite achieving the objective of reducing the polluting agents’ emissions, with~rates of 1.63\% and 3.63\%, respectively, have not been proven to be effective enough~\cite{Long2019}. Consequently, and~given that a~personalized assessment of ecological behavior might help motorists to achieve outstanding fuel consumption results, with~reductions up to 18.4\% \cite{Staubach2014}, providing online DS recommendations seems reasonable. These recommendations must be based on each individual’s driving behavior, and they are intended to re-educate drivers if they follow incorrect driving patterns (e.g.,~aggressive driving) like a~human instructor would~do.

To carry out that task, machine learning (ML) techniques, such as fuzzy logic, have been used to give coaching feedback to the driver about his/her performance~\cite{massoud2019exploring}. Another approach uses artificial neural networks (ANN) to differentiate drivers that are classified among a~plethora of driving behaviors, cycles, and~scenarios, successfully distinguishing between aggressive and defensive behaviors and urban and highway driving~\cite{vaz2014neural}. Several system prototypes for enhancing drivers' braking style employing visual indications have been developed~\cite{delnevo2019combining}. The~authors evaluated the performances of a~variety of ML algorithms while using CAN-bus data jointly with non-invasive ECG sensors and smartwatches. Several~tests have been performed on these prototypes to validate the improvement of drivers' eco-awareness, and~they have yielded promising results. It must be noted that ML-based works on the differences between short-term and long-term DS influence on fuel consumption have been performed with the aim of developing future ADASs, by~means of high-quality models that can accurately predict DS-linked consumption~\cite{Ping2019}, or~by the identification of the critical maneuvers that cause a~rise on the fuel consumption~\cite{Lakshminarayanan2020}. However, those works neither analyze the driver operations on the car commands  (i.e.,~gas pedal, brake pedal, and gear selector) that cause poor fuel economy nor provide concrete recommendations to improve eco-engaged DS.% that can even be extrapolated to electric~cars.

In light of the foregoing information, several limitations regarding eco-driving assessment systems can be identified. (1) Current in-car systems are intended for generic driving recommendations reporting reduced effectiveness. (2) Most personalized driving assessment systems are based on training sessions and fuel-consumption improvement tracking, normally after the driving sessions, achieving low percentages of fuel economy. (3) ML techniques have been successfully applied in order to classify driving styles into several aggressiveness categories; however, the~full potential of these techniques is still unexploited. (4) Most of the existing works in this field do not analyze the handling operations of the driver on the car commands that cause fuel consumption to rise. (5) Providing online DS-based handling recommendations to improve fuel economy is still a~mostly unexplored~path.

%The main challenge addressed in this work is the development of an~eco-driving assessment system, able to provide real-time personalized advice. Moreover, this piece of research investigates the potential of unsupervised neural networks to discover particular driving patterns and examines the underlying causes of different types of non-optimal DSs from the eco-driving viewpoint.
Thus, in~this work, we present the following contributions to the development of an~eco-driving assessment system that is able to provide real-time personalized~advice:
\begin{itemize}[leftmargin=*,labelsep=5.8mm]

    \item New applications of unsupervised neural networks to discover particular driving patterns and analyze the effect of driving patterns in~fuel-consumption.
    
        \item A novel approach for the examination of the underlying causes of different types of non-optimal DSs from the eco-driving viewpoint and analysis of the fuel-economy-compromising command~operations.
    
        \item Personalization of the provided advices when considering the aforementioned points. Those advices comprehend instructions to improve the use of the gas and brake pedals as well as advice on the shift of the selected~gear.
        
         \item Development of a~System-on-Chip with real-time~responsiveness.
         
         \item Improvement in the performance of the already-existing systems, with~expected enhancements in both fuel consumption and emissions ranging from the 9.5\% to the 31.5\%, or~even higher for drivers that are strongly engaged with the system.
\end{itemize}

\subsection{Proposed~Approach}

In this work, we propose an~intelligent system that is able to classify driving behavior, depending on fuel efficiency features and to provide personalized advice according to them. This Advanced Driving Assistance System (ADAS) has been developed while using real-world data from the Uyanik-instrumented car, particularly the data stream from its CAN-bus and the inertial measurement unit (IMU). These~data were collected through driving sessions along a~pre-defined driving path combining urban areas, interurban roads, and~highway~stretches. 

In contrast with the vast majority of works, not only does this proposal classify DS into two or three aggressiveness categories, but~it also analyzes driving behavior by identifying up to five different DSs. This detailed analysis allows for an~insight into the concrete causes of driver-associated high fuel consumption and, consequently, provides personalized DS recommendations to re-educate drivers for eco-friendlier~handling.

The characterization of DS  is performed by means of self-organized maps (SOMs) \cite{Kohonen2001, Kohonen1996}, a~very popular model in the fields of data mining and big data~\cite{Chen2019, Li2020}. It is an~intelligent, unsupervised ML algorithm that is able to automatically group driving behaviors. This ML algorithm has been chosen, because it relies on a~two-dimensional representation of a~high-dimensional complex system, known as a~map, which is suitable for a~qualitative evaluation of multiple driving behavior features. The~car-boarded SOM assessment solution is deployed, utilizing hybrid hardware/software (HW/SW) implementation based on a~field-programmable gate array (FPGA)-based Xilinx ZynQ Programmable System-on-a-Chip (PSoC), which achieves real-time performance~rates.

The aforementioned solution has several advantages when compared to other implementations. The~main one is that DS recommendations are provided in a~natural-language, comprehensible manner to drivers taking their characteristics into account. This enables eco-driving behaviors in motorists that otherwise would not notice eco-driving compromising circumstances.  It is worth remarking that the hardware-based, fully paralleled SOM implementation provides high-speed performance for real-time operation. Besides, it has negligible power consumption and it is compact enough to be boarded in currently marketed cars with almost no~modifications.

The remainder of this paper is organized, as follows. Section~\ref{sec:description} describes the Uyanik dataset, as~well as an~outline of the followed development strategy. In~Section~\ref{sec:DSchar}, the~driving behavior characterization methods for fuel-consumption scenarios are presented, including the selection and obtainment of relevant driving features. Section~\ref{sec:SOM} describes the~application of SOMs for driving behavior classification, while Section~\ref{sec:results} presents experimental results concerning the fuel consumption assessment and emission reduction. Section~\ref{sec:implementation} exposes the implementation and validation of a~hybrid HW/SW PSoC-based fuel-consumption reduction and eco-driving advice system. Section~\ref{sec:concluding} summarizrs concluding remarks.

\section{Outline of the Overall Eco-Driving System~Design}\label{sec:description}

A~system that promotes behavioral adaptations leading to eco-driving is more desirable to encourage drivers to fulfill those requirements, as stated in the preceding sections of this text. For~that purpose, a~real-world data-driven approach was selected. The~data, used to develop a~strategy that identifies eco-driving classes regarding an~individual DS, provide eco-advice based on learning generated from~data.

\subsection{Dataset}
\label{subsec:Dataset}

Several studies have been conducted in the field of intelligent vehicles, with~the spotlight placed on finding out how vehicles, traffic,  the~environment, and~motorists behave as a~set of elements and to determine which sensors are suitable for each situation. For~those reasons, these studies relied on instrumented cars that were fitted with a~variety of different sensors, either those installed by the crew of the study (e.g., inertial measurements units (IMUs), lidars, and~video cameras) or the original ones (e.g., odometer, speedometer, or~the wheels' angular speed sensors). These sensors, jointly~used with loggers of the field buses of the vehicle, provide an~extensive set of raw data that helps researchers to extract meaningful information about individuals' driving and their relationship with the surrounding~environment.

Driving studies, such as NU-Drive~\cite{meiring2015review}, UTDrive~\cite{Angkititrakul2007}, and~Uyanik~\cite{abut2009real}, make use of dedicated instrumented cars, which simplifies data collection and logistics using a~great number of very complex sensors. Such studies are commonly known as non-naturalistic driving studies (non-NDS), which differ from naturalistic driving studies (NDS), in that subjects of study are not driving their own vehicles. In~this context, the~former generally rely on a~very strict experimental control, and~the data they record are more consistent and reproducible, enabling researchers to more easily compare crucial features from different drivers than in NDS, due to motorists driving the same car during~tests.

\begin{figure}[H]
    \centering
    \includegraphics[width=0.9\textwidth]{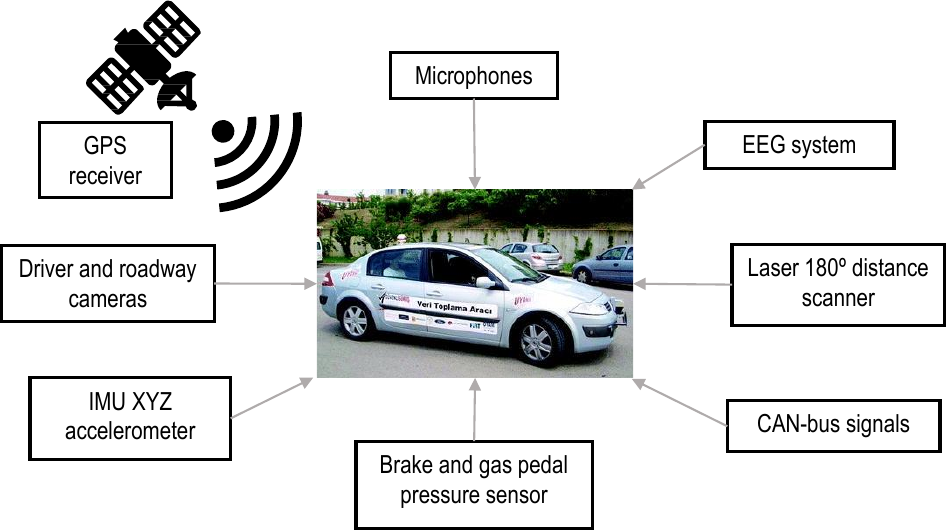}
    \caption{\hl{Data-acquisition} systems and sensors installed in the Uyanik car~\cite{abut2009real}.}
    \label{fig:figinstru}
\end{figure}

In this work, we used the Uyanik-instrumented car dataset~\cite{abut2009real}. Uyanik~\cite{abut2009real, Abut2007} is a~Renault Mégane sedan, fitted with a~reinforced front bumper, a~high-power battery, a~1500 W DC-AC power converter, a~CAN-bus output socket, an~instrument bench at the navigator's seat, and power and signaling rewiring (see Figure~\ref{fig:figinstru}). The~complete dataset that was compiled by the vehicle on each test drive comprises three channels of uncompressed video from the left and right sides of the driver and the road ahead. It also includes three audio recordings, GPS, and~CAN-bus readings, including vehicle speed (VS), engine RPM (ERPM), steering wheel angle (SWA), and~brake pedal status (pressed or idle) (see Table~\ref{tab:variables}). Gas pedal engagement percent (PGP) is sampled at either 10 or 32 Hz, whereas brake pedal and gas pedal pressure sensor readings (BP and GP, respectively) are sampled at the same CAN-bus rate. Finally, a~laser distance measuring device was fitted in the front bumper jointly with an~IMU XYZ Acceleration measuring sensor set-up and an~electroencephalogram (EEG) monitor. In~this study, all of the signals were handled jointly, which requires a~re-sampling of the data streams to the highest frequency of 32~Hz.

Video channels were collected by means of a~digital video recorder, audio by a~data acquisition system, and~digital data by a~merge of all the signals with RS-232 and USB buses into a~laptop computer running custom software that was developed by the Technical University of Istanbul~\cite{abut2009real}. It is worth noting that the audio and video feeds, as well as the digital data, were properly~synchronized.

%MDPI: We insert it close to where it is first mentioned in the text. please confirm and check below.
%Author: Change confirmed
\unskip

\begin{table}[H]
\caption{Significative variables of the Uyanik dataset~\cite{abut2009real}.}
\label{tab:variables}
\centering
\begin{tabular}{ll}
\toprule
\textbf{Features}                          & \textbf{Signals (Time Series)}       \\ \midrule
\multirow{6}{*}{CAN bus\vspace{-5pt}}          & Steering wheel angle (SWA)   \\ \cmidrule(lr){2-2} 
                                  & Steering wheel speed (SWS)   \\ \cmidrule(lr){2-2} 
                                  & Vehicle speed (VS)         \\ \cmidrule(lr){2-2} 
                                  & Percent gas pedal (PGP)     \\ \cmidrule(lr){2-2} 
                                  & Engine RPM (ERPM)           \\ \midrule
\multirow{2}{*}{Pressure sensors\vspace{-5pt}
} & Brake pedal pressure (BP)  \\ \cmidrule(lr){2-2} 
                                  & Gas pedal pressure  (GP)    \\ \midrule
\multirow{8}{*}{IMU unit\vspace{-5pt}}         & Roll rate    (RR)           \\ \cmidrule(lr){2-2} 
                                  & Pitch rate  (PR)            \\ \cmidrule(lr){2-2} 
                                  & Yaw rate   (YR)             \\ \cmidrule(lr){2-2} 
                                  & X axis accelerometer (XACC)  \\ \cmidrule(lr){2-2} 
                                  & Y axis accelerometer (YACC) \\ \cmidrule(lr){2-2} 
                                  & Z axis accelerometer  (ZACC)\\ \midrule
Laser                             & Distance to obstacle (d\_90)\\ \midrule
GPS                            & Coordinates (GPS) \\ \bottomrule
\end{tabular}
\end{table}

\subsection{Development~Strategy}

\begin{figure}[H]
    \centering
    \includegraphics[width=1\textwidth]{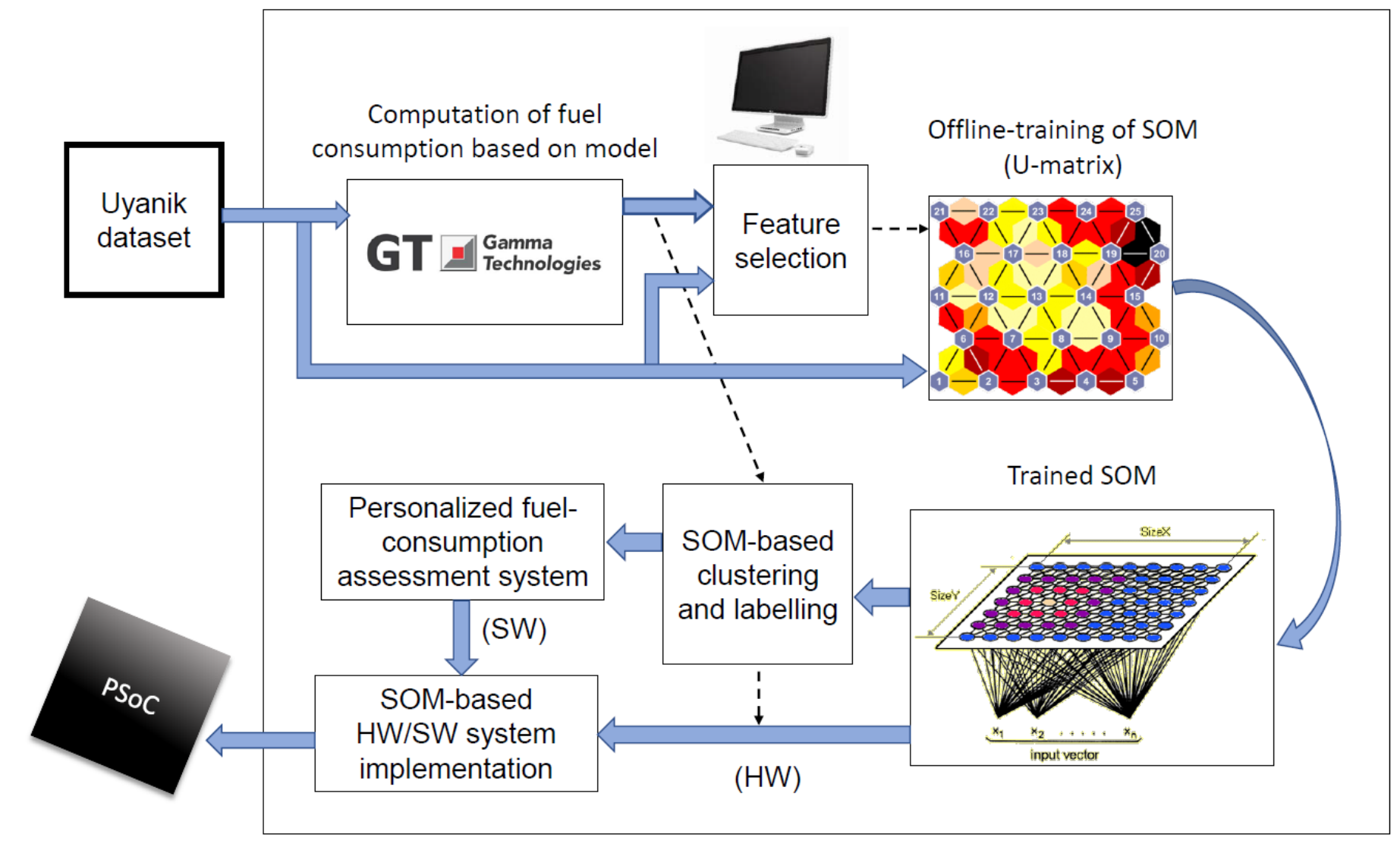}
    \caption{Offline sequence of  tasks involved in the design and development of an~self-organized map (SOM)-based intelligent system for fuel consumption assessment. The~dotted arrows indicate that the simulated fuel consumption data are also used to label the SOM-based clustering for verification purposes and to elaborate the hardware (HW) implementation of the~SOM.}
    \label{fig:figbloq}
\end{figure}

The development strategy that was used in this work combines several algorithms and tools in a~multi-stage fashion. Five stages can be clearly identified from the raw data itself until  %please check, incorrect use of "until" suspected
a hybrid HW/SW integrated system to recommend drivers about changing their DS is obtained (see Figure~\ref{fig:figbloq}).

\subsubsection{Feature~Selection}
This initial stage is one of the most work-intensive, and~it comprises the use of several resources from two different sources: the Uyanik car dataset and GT-Suite simulation tool~\cite{Gtisoft2017}. The~real-world driving data were input into the GT-Suite simulator, where a~realistic model of the car was emulated. This simulation allowed for us to obtain the fuel consumption flows that the Uyanik dataset lacked. Afterwards, the~most relevant features as well as the optimal data window size were selected. The~features exhibiting the strongest relationship with fuel consumption were the mean values of the percent gas pedal (PGP),  the~engine RPM (ERPM),  the~gas pedal pressure (GP), and~the variance of the positive acceleration in the X axis (Pos XACC), all from the Uyanik dataset. Regarding window size, it~was found that a~256-sample window size (that is to say, 8 s of data at a~sample rate of 32 Hz) with an~overlapping of 50\% returned the most appropriate~results.

\subsubsection{Offline-Training of~SOM}
 This step consists of mapping a~four-dimensional input space: mean (PGP, ERPM, GP) and variance (pos XACC) on a~two-dimensional grid of neurons (i.e.,~the SOM) that keeps the principal features of the inputs. Thus, given the set of input samples, the~weights of the neurons are sequentially updated in a~competitive and collaborative way with the aim of minimizing the distance between each input sample and the corresponding winning neuron (i.e.,~the best matching unit, BMU), until~a~stop criterion is reached. Different SOM topologies were explored, which ranged from 10 $\times$ 10 neuron maps ($M=100$) to 15 $\times$ 15 neurons ($M=225$), by~using the Matlab Neural Network Clustering App~\cite{Matlab2020}. The~most robust and consistent results were obtained using 11 $\times$ 11 maps ($M=121$). This training process, as detailed in Section~\ref{subsec:SOM_train}, is completely unsupervised, that is to say, the~GT-Suite data were not used during the offline-training~step.

\subsubsection{Clustering and Labeling of Trained~SOM}
\label{subsub:clusSOM}
This step consists of performing both a~quantitative and qualitative analysis of the trained SOM. The~former is based on a~visual inspection of the so-called U-matrix (i.e.,~a~unified distance matrix), which provides a~visual representation of the distances between neighboring neurons by using a~color scale. The~latter evaluates the U-matrix mathematically. Thus, this matrix is useful for identifying clusters both graphically and numerically (see Section~\ref{subsec:SOM_clustering}). This tool helps to see the cluster structure of the map: high values of the U-matrix indicate cluster border, while uniform areas of low values can be identified as potential clusters. The~quantitative evaluation of the SOM was performed by means of the CIS SOM Toolbox for Matlab~\cite{Vesanto1999}. According to the identified clusters, as~described in Section~\ref{sec:clustering_results}, several groups were identified and labeled according to the mean fuel consumption that was obtained by simulation with~GT-Suite.

\subsubsection{Development of Driver~Advice}
With the properly labeled clusters, meaningful three-dimensional plots of the selected features were analyzed with the aim of discovering the aspects that each DS group can improve. After~that, several fuel-consumption-compromising circumstances were identified (see Section~\ref{sec:grouping_results}). Next,~with~those pieces of information, concrete actions were developed in order to provide personalized advice to the drivers. These actions suggest how drivers can modify their DS to achieve better consumption rates, with~a measurable effect on drivers' eco-consciousness and eco-driving. Finally, the experimental results of particular Uyanik drivers were~analyzed. 

\subsubsection{PSoC-Based HW/SW~Implementation}
Finally, the~personalized fuel consumption assessment system was developed and implemented on an~SoPC by means of the VHDL hardware description language and the Xilinx Vivado 2018.1 design suite~\cite{Vivado2018}. The~whole system architecture is composed of an~HW partition, an~SW partition, and~an internal communication interface (see Section~\ref{sec:implementation}). A~fully parallel, high-performance, SOM accelerator core was deployed in the FPGA part of the device (i.e.,~the HW partition). This HW partition is intended to achieve real-time performance rates, so as to perform an~online assessment of DS. The~SW partition, which interfaces with the automobile's buses, performs I/O exchanges, computes the features' windows, and~gives advice to the driver, depending on the SOM-based clustering results. The~on-chip HW/SW communication is performed by means of standard Advanced eXtensible Interfaces (AXI).

\section{Driving Behavior Characterization for Fuel-Consumption~Scenarios}
\label{sec:DSchar}

Energy~consumption and carbon dioxide emissions of passenger cars are affected by a~combination of human, environmental, and~technological factors, according to a~recent report of the Joint Research Centre of the European Union~\cite{zacharof2016review}. Human factors refer to driving behavior, that is to say, the~driving patterns that an~individual driver or a~group of drivers follows, such as acceleration, mean speed, and~preferred engine gear. The~main environmental factors include both weather conditions (i.e.,~ambient temperature, rain, and~wind) and actual characteristics of the road (i.e.,~morphology, surface quality, and~traffic conditions), while technological factors refer to the vehicle type and its~characteristics.

In this work, we focused on the consequences of the DS on fuel consumption, so the human factor had to be isolated as much as possible from the remaining factors~\cite{Tanvir2018,Tanvir2019}.  With~the aim of fulfilling the above requirement, we considered a~group of drivers exhibiting different driving behaviors while driving the same car, along the same route, and~in similar environmental conditions. It is worth noting that the latter factor, mainly traffic conditions, is the most difficult feature to reproduce in live~traffic.

\subsection{Selection of Relevant~Features}
\label{sec:features}

The~dataset used in our experiments was collected using an~instrumented car traveling a~fixed route around the city of Istanbul, as already introduced in Section~\ref{subsec:Dataset}. The~route is little over 25 km and lasts about 40 min, depending on weather and traffic conditions. It includes different types of road sections: city, very busy city, highway, and~a university campus. With~the aim of minimizing the impact of environmental variations, all of the selected driving sessions were conducted in a~short period of time, from~August to October, and~during the central part of the day, between~11 a.m. and 4 p.m. The~driver population was composed of 20 drivers, 17 male and three female, whose ages ranged from 21 to 61. This is a~reduced subset of a~more comprehensive data collection (i.e.,~about 100 drivers and a~single trip per driver) provided by the “Drive-Safe Consortium” \cite{abut2009real}.

Because instant fuel consumption was not available within the dataset, we developed a~model of the Uyanik car and used the GT-Suite tool to obtain fuel consumption data during the driving sessions~\cite{asfoor2014use}. Afterwards, we computed two types of features: mean values and variances of the Uyanik signals. The~whole set of time series, more than 30 independent signals, was evaluated, including CAN-bus data, pedal pressure sensors, a~laser scanner, and~IMU unit readings. In~particular, the~treatment of the X-axis acceleration variables was divided into two parts, positive and negative values, since they have different consequences on fuel consumption. In~fact, negative instantaneous values are associated with zero~consumption.

Subsequently, the~features that provide the highest relationship with fuel consumption were selected, while the irrelevant or redundant features were discarded. We computed both the Pearson correlation coefficients (PCCs) and the \emph{p}-values of every feature. The~former provides a~measure of the relevance of each feature, while the latter is used for testing the hypothesis of no correlation (i.e.,~the probability of obtaining a~correlation as large as the observed value by random chance, when the true correlation is zero). The~features were computed over 8 s windows (i.e.,~256 samples) with a~4 s shift. That is to say, the~overlapping between consecutive windows is 4 s (i.e.,~128 samples). The~format of the windows was selected by exhaustively analyzing the consequences of both the window size and the shift on the PCC of the most relevant features. Table~\ref{tab:PCC_mean_var} summarizes the set of low level signals (i.e.,~time~series) that exhibit the strongest correlation with fuel consumption. Moreover, the \emph{p}-values are less than 0.0001 for almost all of the features included in this table, thus guaranteeing the reliability of the correlations. The~exceptions are the mean and variance of the negative X-axis acceleration, whose \emph{p}-values are close to 0.05. These features were not selected because of their low~PCCs.

\begin{table}[H]
%\renewcommand{\arraystretch}{1.5}
% if using array.sty, it might be a~good idea to tweak the value of
%\extrarowheight
\caption{Driving behavior signals and Pearson correlation coefficients (PCCs) of fuel consumption with relevant features. Mean values and variances are computed using 8 s analysis~windows.}
\label{tab:PCC_mean_var}
\centering
\begin{tabular}{p{3cm} p{4cm} c c}%p{2cm} p{2cm}} %{lllll}
\toprule
\textbf{Measurement Units} & \textbf{Signals (Time Series)} & \textbf{PCC: Mean} & \textbf{PCC: Variance}\\
\midrule
\multirow{4}{2cm}{CAN-bus} & Vehicle speed (VS) & \textbf{\hl{0.59}} & 0.15\\ \cmidrule(lr){2-4}
& Percent gas pedal (PGP)& \textbf{\hl{0.63}} & \textbf{\hl{0.58}}\\ \cmidrule(lr){2-4}
& Engine RPM (ERPM) & \textbf{0.66} & 0.18\\
\midrule
\multirow{2}{2cm}{Pressure sensors} &  Brake pedal pressure (BP) & $-$0.35 & $-$0.23\\ \cmidrule(lr){2-4}
& Gas pedal pressure (GP) & \textbf{0.52} & 0.20\\
\midrule
\multirow{4}{5cm}{IMU unit\vspace{-5pt}
}
& Positive X axis acceleration (Pos XACC) & 0.32 & 0.25\\ \cmidrule(lr){2-4}
& Negative X axis acceleration (Neg XACC) & $-$0.17 & $-$0.11\\
\bottomrule

\multicolumn{4}{p{12cm}}{\footnotesize \justifyorcenter{The boldface PCCs correspond to the strongest correlations.}}
\end{tabular}
\end{table}
%MDPI: Is the bold necessary? If yes, please add explain for bold. Please confirm and check below.
%Author: We considered it is necessary because it helps to better understand the table.
\vspace{-15pt}

The features that present the strongest correlations with fuel consumption are remarked in bold in Table~\ref{tab:PCC_mean_var}. Four mean values (i.e.,~VS, PGP, ERPM, and~GP) and the PGP variance have a~strong positive correlation with fuel consumption, while the positive X-axis acceleration presents moderate correlations, as~can be seen. On~the other hand, BP and negative X-axis acceleration, both mean and variance, exhibit negative correlation coefficients. This means that an~increase in BP or in X-axis deceleration is associated with a~decrease in fuel consumption. Although~these features are meaningful concerning driving behavior analysis, their correlations with fuel consumption are rather~weak.

Afterwards, we chose the most fuel-demanding sections of the Uyanik route, those that ran through highway and motorway, in order to~develop the assessment system. Moreover, sections with traffic jams and slow traffic (i.e.,~mean speed below 60 km/h) were discarded with the aim of avoiding outliers during the training process. After~limiting the type of road, the~PCCs, as presented in Table~\ref{tab:PCC_mean_var}, varied~slightly. The~most noticeable changes were a~moderate reduction of the fuel consumption correlations with VS and a~remarkable increase in the fuel consumption correlations with the mean and variance of positive X-axis acceleration. In~view of these results, the~latter features were also taken into account in a~preliminary round of SOM training experiments. Thus, the~following variables were selected as candidate features: mean values \{VS, PGP, ERPM, GP, Pos XACC\} and variances \{PGP,~Pos~XACC\}. A~comprehensive series of training experiments revealed that a~reduced subset of only four features is able to model the relationship between fuel consumption and driving behavior in a~very satisfactory way. These features were mean PGP, mean ERPM, mean GP, and~Pos XACC~variance.

\subsection{Fuel-Consumption Obtainment by~Simulation}

An~important step of this work is the obtainment of a~meaningful set of fuel consumption data, as mentioned in precedent paragraphs. This step was found to be needed, since the Uyanik dataset lacked the engine's engine control unit (ECU) data regarding fuel injection or intake~airflow. 

\begin{figure}[H]
    \centering
    \includegraphics[width=0.9\textwidth]{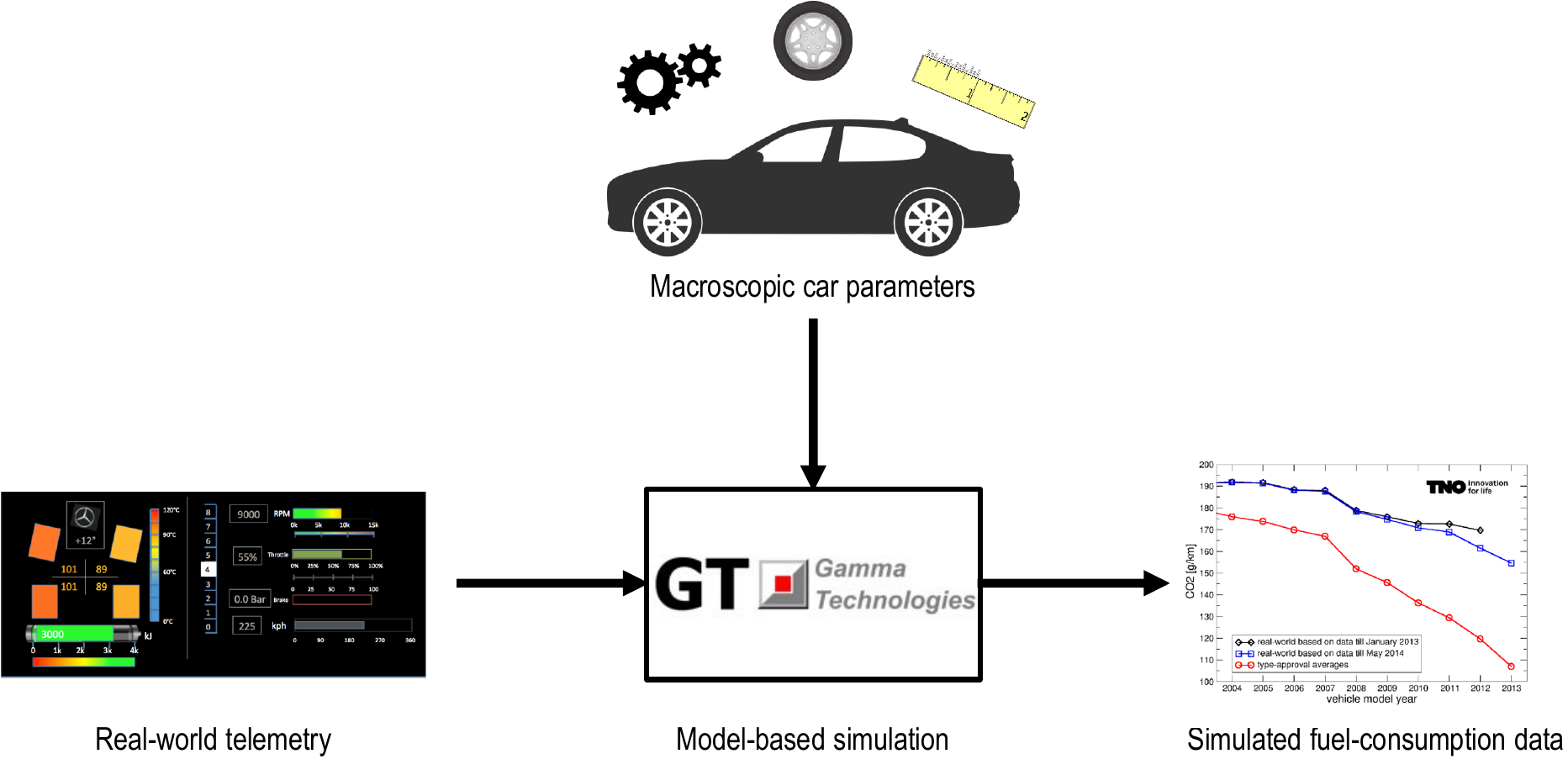}
    \caption{\hl{Flow of real}-world telemetry-based fuel consumption simulation. It has macroscopic car parameters (gear ratios, tyre dimensions, and~wheelbase) and telemetry (gas pedal, brake pedal, speed, selected gear, and~accelerations) as inputs. The~model returns the simulated fuel flow as~output.}
    \label{fig:figsimuflow}
\end{figure}

Several alternatives were studied in order to obtain and measure fuel control unit data, finally choosing a~simulation environment. We selected the Gamma Technologies GT-Suite environment~\cite{Gtisoft2017}, since~it does not only allows element-by-element simulation of mechanical systems, but~also enables users to run macroscopic approximated models of complete automobiles. Thus, while the former requires an~exact parameterization of each mechanical element and link of the engine, the~latter allows us to fit a~pre-elaborated model based on telemetry (such as speed, acceleration, brake, or~selected gear) as well as on car manufacturer information, such as gear ratios, tire dimensions, or~wheelbase (see Figure~\ref{fig:figsimuflow}).

%MDPI: Please replace with a sharper image.
%Author: Sharpness of the image improved. On the other hand, we checked the definition of the image when printed and it's correct

Finally, the~gear ratios were computed while using data about RPM and vehicle speed, available for each instant. We computed the speed/rpm ratio sample-by-sample and matched it with each gear's ratio. When computed ratios did not match with any of the gear ratios, we assumed that the driver was operating the clutch~pedal. 

\begin{figure}[H]
    \centering
    \includegraphics[width=0.90\textwidth]{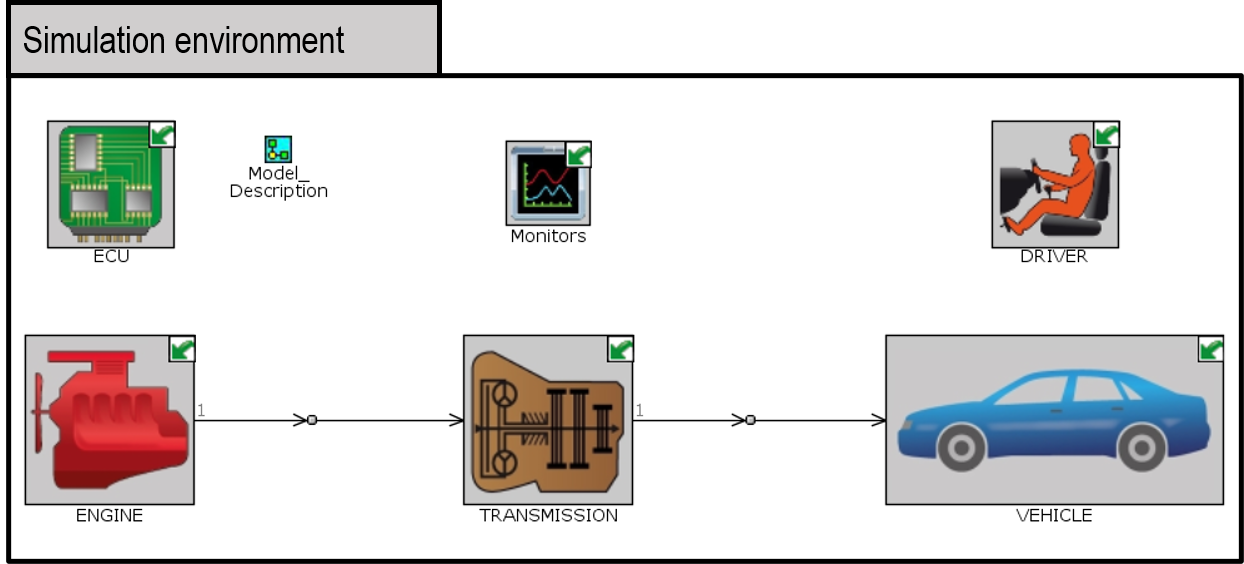}
    \caption{Block diagram of real-world telemetry-based fuel consumption simulation of Uyanik Renault Mégane 1.5 dCi Sedan 74 kW. It has macroscopic car parameters (gear ratios, tyre  dimensions, and~wheelbase) and telemetry (gas pedal, brake pedal, speed, selected gear, and~accelerations) as inputs. The~model returns the simulated fuel flow as~output.}
    \label{fig:entornosimu}
\end{figure}

Once the car parameters were successfully extracted, we elaborated on the model that is displayed in Figure~\ref{fig:entornosimu}. In~this model, four main elements can be identified for the car itself, the~vehicle, transmission, engine, and~ECU blocks, while the driver is modeled by another one%another what? please specify
. These blocks contain the characteristic parameters of their corresponding real-world~counterparts.
\begin{itemize}[leftmargin=*,labelsep=5.8mm]
    \item Vehicle comprises data regarding car wheelbase, wheel radius, friction coefficients, aerodynamics, weight, inertia, and~final transmission ratio.
    \item Transmission incorporates individual ratios for each of the user-selectable gears, as~well as clutch~parameters.
    \item Engine consists of parameters such as engine displacement, engine type (4-stroke or 2-stroke), minimum operation speed, or~fuel characteristics.
    \item ECU controls the maximum engine RPM, idle speed, and~fuel injection cutoff and restart points.
    \item Driver wraps the telemetry data related to the handling of the car, such as selected gear, accelerator~pedal state, brake pedal state, clutch, and~desired speed.
\end{itemize}
%MDPI: Is the italic necessary? Please confirm and check below.

\begin{figure}[H]
    \centering
    \includegraphics[width=0.95\textwidth]{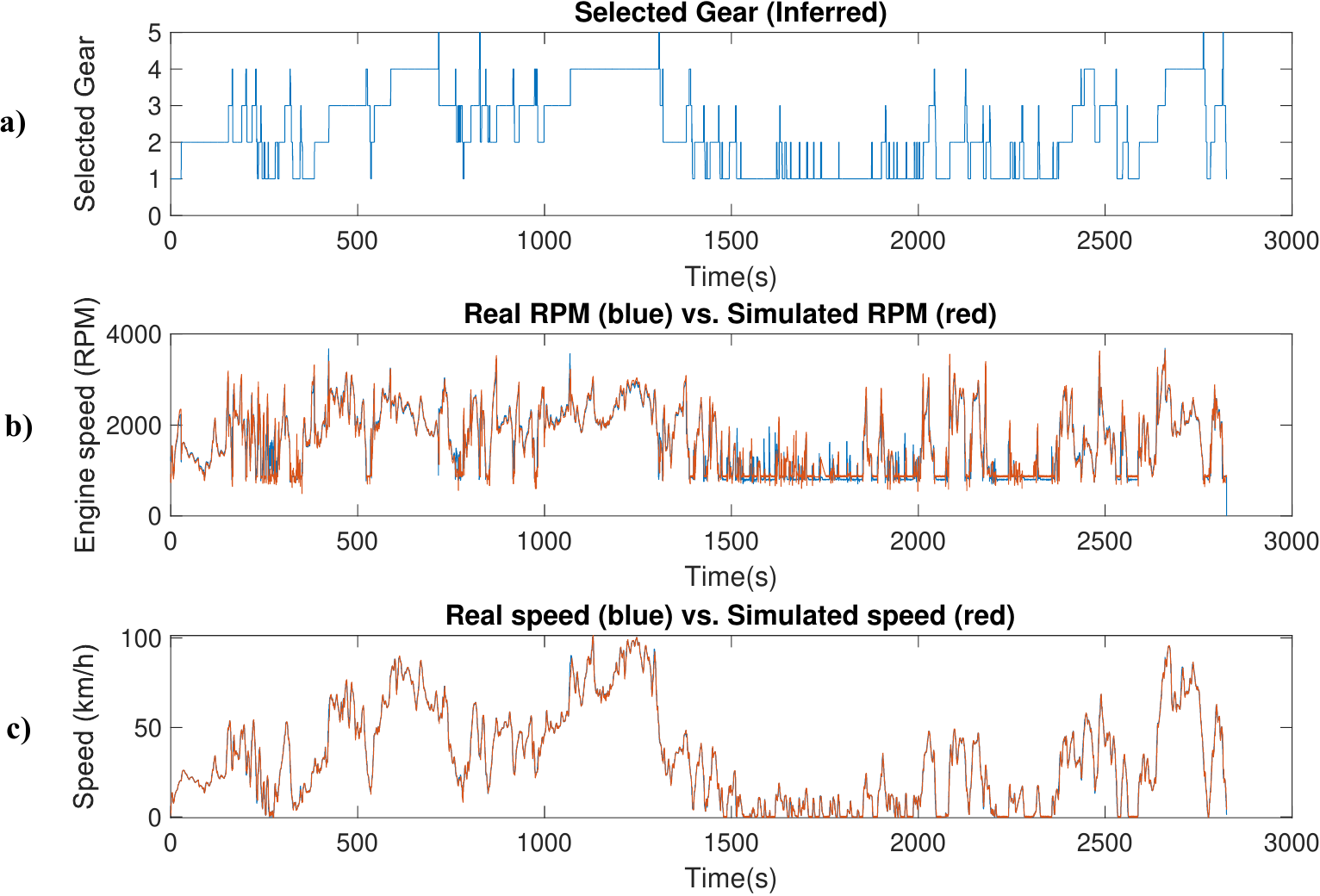}
    \caption{Comparison of measured data vs. simulation results. (\textbf{a})   The inferred gear considering the computed rpm/speed ratios. (\textbf{b}) The measured revolutions per minute (RPM) of the vehicle vs. the RPM simulated by the model. (\textbf{c})  The measured speed of the vehicle vs. the speed simulated by the~model.}
    \label{fig:simumatlab}
\end{figure}

Several checks were performed on the simulation model in order to verify that the returned results provide an~acceptable emulation of the real car performance. Thus, given a~set of selected gears, as displayed~in Figure~\ref{fig:simumatlab}a, the~application of the driver operation of the accelerator and brake pedals, along with the dynamics of the car restricted to a~set of measured accelerations, brings out the simulated RPM and vehicle speed red curves of Figure~\ref{fig:simumatlab}b,c, respectively. As~can be seen, these red curves are almost totally overlapped with the blue ones, which represent the real world-collected data, with~relative errors of 1.83\% for RPM and of 0.44\% for speed. These low relative errors mean that the simulation faithfully emulates the real car behavior and, consequently, that the returned fuel consumption simulated data is useful for carrying out estimations in order to verify the proposed SOM-based models and extracting~conclusions.

\section{Self-Organizing Maps Applied to Driving-Behavior~Classification}
\label{sec:SOM}

The SOM is a~particular type of ANN suitable for clustering and visualization of complex multi-dimensional data~\cite{Kohonen1996,Kohonen2001, Miljkovic2017}. It defines a~mapping, or~projection, from~a set of high-dimensional input data onto a~regular low-dimensional discrete grid. This grid, which is known as a~feature map, preserves the principal features of the input data. Unlike conventional feed-forward ANNs, which are generally trained using the supervised back-propagation learning algorithm, SOMs are trained through an~unsupervised strategy; that is to say, in~an SOM, there are no known target outputs that are associated with each input sample. On~the contrary, during~the training phase, SOMs process a~collection of data, only~input data, in order to~discover unknown clusters hidden in the~data.

\begin{figure}[H]
    \centering
    \includegraphics [width=0.7\textwidth]{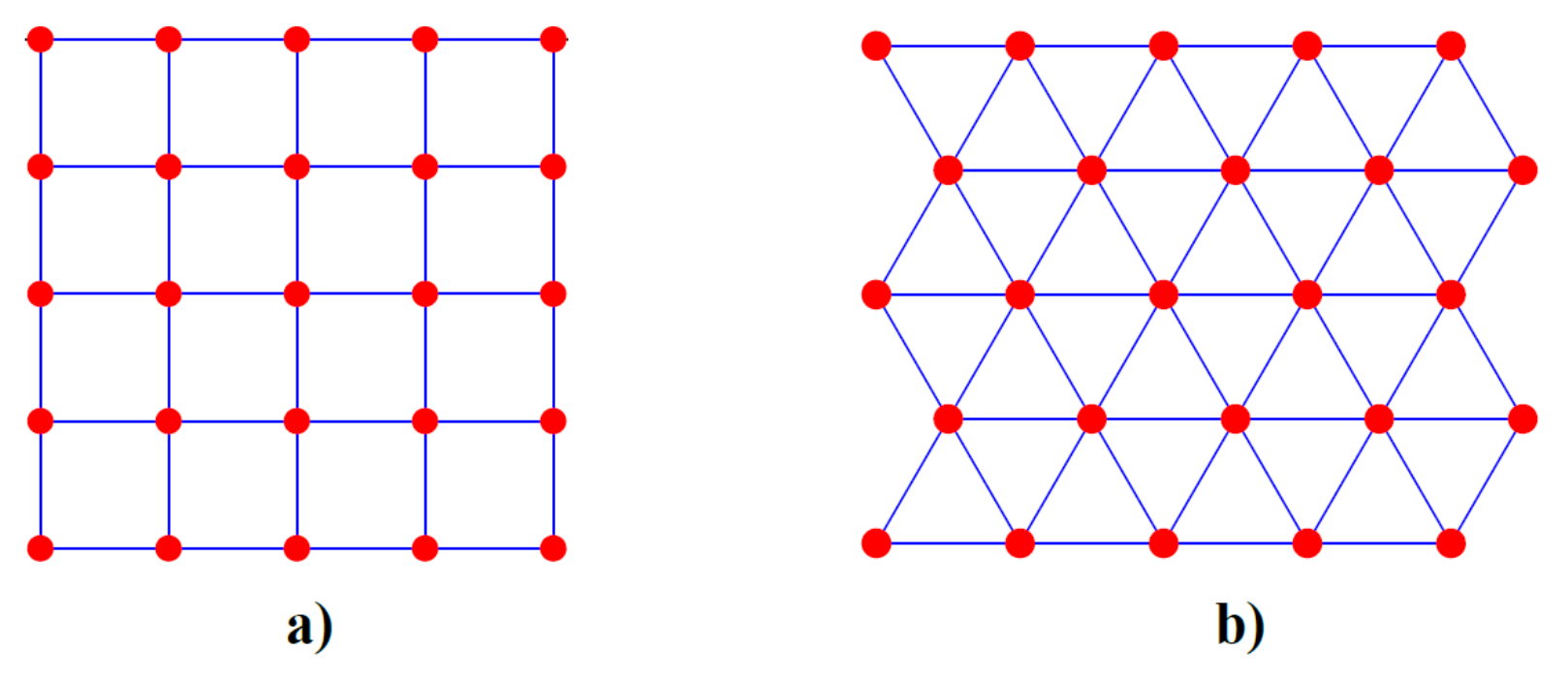}
    \caption{Typical SOM topologies: a~rectangular output grid (\textbf{a})  and a~hexagonal output grid (\textbf{b}).}
    \label{fig:grid_hex}
\end{figure}

The architecture of an~SOM consists of a~single layer neural network with neurons set along a~regular grid: the output layer. Each input to the SOM is fully connected to every neuron in the output layer. Figure~\ref{fig:grid_hex} depicts two typical two-dimensional output layers with $M=25$ neurons set along a~rectangular grid (a) and a~hexagonal grid (b). Although~most of the SOMs are based on a~two-dimensional grid, many applications also use three or more dimensional~spaces.

\begin{figure}[H]
    \centering
    \includegraphics [width=0.6\textwidth]{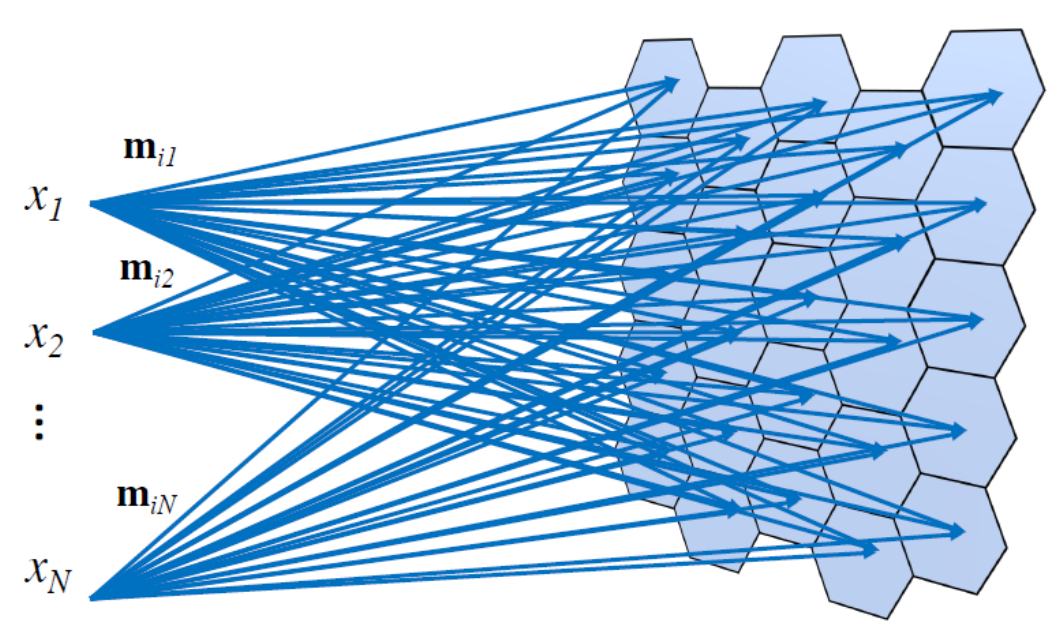}
    \caption{Structure of an~SOM with $N$ inputs, $\mathbf{x} = (x_{1},x_{2},\cdots,x_{N})$, and~$M$ = 25 output neurons distributed into a~5 $\times$ 5 two-dimensional hexagonal~grid.}
    \label{fig:SOM_structure}
\end{figure}

Each neuron in the output layer has a~double representation: an~$N$-dimensional vector $\mathbf{m}_{i}$, known~as the weight vector, and~its position in the grid. The~number of components of the vector is equal to the number of input features $N$. Figure~\ref{fig:SOM_structure} shows the structure of the SOM that was used in this work, and~it is based on a~two-dimensional hexagonal~grid.

\begin{equation}
\mathbf{m}_{i}=(m_{i1},m_{i2},\cdots,m_{iN}), 1 \leq i \leq M.
    \label{eq:weight_vector}
\end{equation}

Clustering a~dataset by means of an~SOM paradigm is carried out using a~two-level approach: first, the~SOM is trained;  afterwards, the~prototype vectors of the SOM are clustered. The~advantage of using this approach, instead of clustering the data directly, is that the computational load decreases considerably, which makes it suitable for analyzing different pre-processing and initialization strategies in a~short time. In~addition, a~two-level approach is less sensitive to noise than a~single-level strategy. Obviously, this solution is only valid if the clusters found using the SOM are representative of the original data~\cite{Vesanto2000}.

\unskip
\unskip

\subsection{Training Self-Organizing~Maps}
\label{subsec:SOM_train}
First, an~initial weight is assigned to each neuron connection. There are simple initialization approaches, such as using random numbers or using input samples randomly selected from the dataset. Although~sophisticated algorithms that are based on data analysis (e.g., principal component analysis (PCA)) can also be used, it was observed that random initialization performed rather well for non-linear datasets~\cite{Akinduko2012}.

Thus, in~each training step, one input sample $\mathbf{x}^{k} = (x_{1}^{k},x_{2}^{k},\cdots,x_{N}^{k})$, $1 \leq k \leq K$, from~the dataset is chosen randomly, and~the distances between this sample and all of the neuron weights of the SOM are computed. The~most popular distance measure in real applications is the Euclidean distance $\left\Vert \cdot \right\Vert$.
\begin{equation}
\left\Vert \mathbf{x}^{k} - \mathbf{m}_{i} \right\Vert^{2} = \sum_{j=1}^{N} (\mathbf{x}_{j}^{k} - \mathbf{m}_{ij})^{2}.
\label{eq:Euclidean}
\end{equation}

The output neuron whose weight vector is closest to the $k$-th input sample, according to Equation~\eqref{eq:Euclidean}, is the best matching unit (BMU) or the winner neuron, which is usually denoted by $c$.
\begin{equation}
\left\Vert \mathbf{x}^{k} - \mathbf{m}_{c}^{k} \right\Vert = \mathop{min}_{i}\left\Vert \mathbf{x}^{k} - \mathbf{m}_{i} \right\Vert.
\label{eq:BMU}
\end{equation}

The BMU is used to update the weight vectors of the SOM. In~this process, the~BMU and its neighbors are moved towards the $k$-th input sample, bringing them closer. For~each neuron of the SOM, the~weight vector is updated, as follows:
\begin{equation}
\mathbf{m}_{i}(n+1) = \mathbf{m}_{i}(n) + \alpha(n) h_{ci}(n) \left\Vert \mathbf{x}^{k}(n) - \mathbf{m}_{i}(n) \right\Vert
\label{eq:update_rule}
\end{equation}
where $n$ denotes the iteration step, $\mathbf{x}^{k}(n)$ is an~input sample randomly selected from the training dataset at iteration $n$, $h_{ci}(n)$ is a~neighborhood function or kernel around the BMU, and~$\alpha(n)$ is the learning rate. Both $\alpha(n)$ and $h_{ci}(n)$ are decreasing functions approaching zero with each iteration in~order to guarantee the convergence and stability of the training process. The~neighborhood function specifies how much the $i$-th neuron has to move toward the input sample at iteration step $n$. It is a~radial basis function, usually a~Gaussian function that is centered at the BMU:
\begin{equation}
h_{ci}(n) = exp\left(-\frac{\left\Vert \mathbf{m}_{c}(n) - \mathbf{m}_{i}(n) \right\Vert^{2}}{2\sigma^2(n)}\right).
\label{eq:kernel}
\end{equation}

Equation \eqref{eq:kernel} defines the region of influence of the current input sample, with~$\sigma^2(n)$ being the neighborhood~radius.

Concerning the learning rate in Equation \eqref{eq:update_rule}, different functions have been proposed, such as linear or exponential functions. In~this work, a~decaying exponential, with~initial learning rate $\alpha_{0}$, has~been selected:
\begin{equation}
\alpha(n) = \alpha_{0} e^{\left(-\frac{n}{T}\right)}
\label{eq:learning_rate}
\end{equation}
where $T$ is the number of iterations or training length. It is worth noting that Equation \eqref{eq:update_rule} is also suitable for the online training of SOM by substituting $n$ by $t$ (i.e.,~discrete time).

In sum, the~sequential training of SOMs involves the following~steps:
\begin{enumerate}[leftmargin=*,labelsep=4.9mm]
    \item initialization. Initial weights are randomly selected from the dataset.
    \item Competition. For~a randomly selected input sample, all of the neurons in the output layer compete with each other to be the BMU (Equation \eqref{eq:BMU}). The~neuron that is closer to the input sample is the winner.
    \item Cooperation. The~BMU also excites the neurons in its topological neighborhood. This cooperative process decays as neurons are further away from the winning neuron (Equation \eqref{eq:kernel}).  
    \item Adaptation. The~BMU and its neighboring neurons are pulled closer to the input sample. For~each neuron in the SOM, the~weight vector is updated according to Equation \eqref{eq:update_rule}.
\end{enumerate}

After initialization, the~remainder training steps are repeated until a~stop criterion is~achieved.

\subsection{Clustering of Self-Organizing~Maps}
\label{subsec:SOM_clustering}

The above unsupervised learning algorithm preserves data topology; that is to say, samples that are close together in the high-dimensional input space have close positions in the map. After~training, the~SOM provides a~nonlinear mapping of the dataset onto a~two-dimensional grid that allows for identifying groups of samples with similar characteristics (i.e.,~clusters) by taking all features into account simultaneously.

\begin{figure}[H]
    \centering
    \includegraphics [width=1\textwidth]{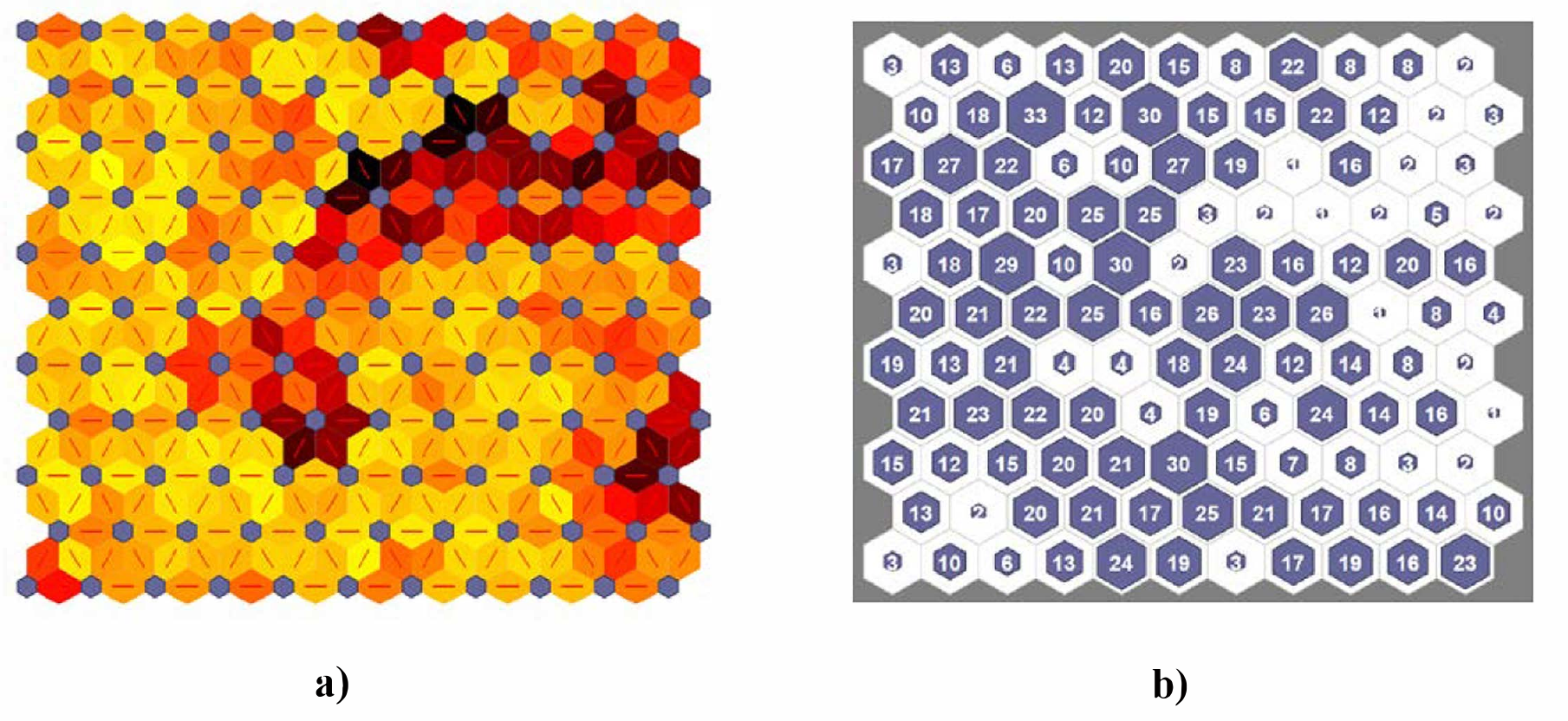}
    \caption{SOM organized into an~11 $\times$ 11 neuron grid. (\textbf{a}) Neighbor weight distances. The~blue hexagons represent the output neurons, while the red lines are neuron connections. Darker colors represent larger distances between neighboring neurons, and~lighter colors represent smaller ones. (\textbf{b})~Sample hits. This image shows how many training samples are associated with each~neuron.}
    \label{fig:Umatrix_and_hits}
\end{figure}

The unified distance matrix (U-matrix) provides the distances of the weight vectors to each of its immediate neighbors in the grid. It can be used with the aim of displaying the distance structures, while using a~color scale in the two-dimensional array of neurons, maintaining the topology, and~allowing for the identification of the clusters, boundaries, and~representative neurons. See, for~example,  Figure~\ref{fig:Umatrix_and_hits}a, where the U-matrix of an~11 $\times$ 11 hexagonal SOM topology is shown. The~U-matrix is not only a~useful visualization tool but also a~powerful analysis tool suitable for mathematically identifying clusters. Another useful visualization method consists in displaying the number of hits in each neuron of the map. This information can also be applied in clustering the SOM using low-hit neurons to locate cluster borders (see Figure~\ref{fig:Umatrix_and_hits}b). In~this work, the~U-matrix based clustering algorithm was used~\cite{Vesanto1999}.

\subsection{SOM-Based Drivers Grouping Regarding~Fuel-Consumption}
\label{sec:clustering_results}

As the number of neurons in the map ($M= 11\times11 = 121$) is less than the number of samples ($K=1717$), most of the neurons in the map are the BMU or hit of several samples in the dataset (see~Figure~\ref{fig:Umatrix_and_hits}b). As~can be seen, there are neurons across the map with 4 or fewer hits, which match the regions with dark neuron connections in Figure~\ref{fig:Umatrix_and_hits}a. These units could be considered to be interpolating neurons, smoothing the transitions between~clusters.

\subsubsection{SOM Classification~Results}
\label{sec:grouping_results}

The above visualizations of the trained SOM can only be used to obtain qualitative information concerning driving behavior. Interesting groups of neurons (i.e.,~clusters) must be identified and labeled in order to~develop meaningful quantitative descriptions of driving data, suitable for a~real-time fuel consumption assessment. Although~the clustering of the SOM can be performed by means of any unsupervised clustering method, such as K-means or hierarchical clustering, the U-matrix method was~used in~this work.

\paragraph{Three-Cluster Grouping}

\begin{figure}[H]
    \centering
    \includegraphics[width=0.7\textwidth]{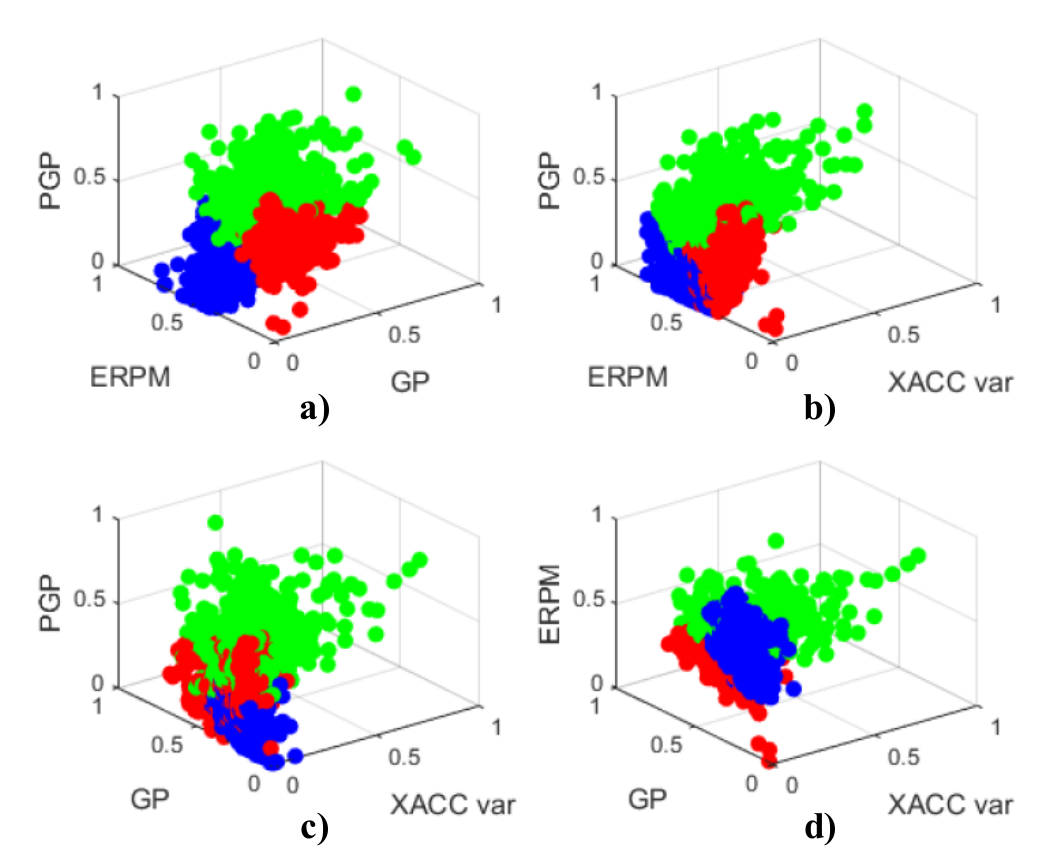}
    \caption{\hl{Three-dimensional} views of the three-cluster fuel consumption classification results. The~clusters were labeled as Very low (red), Low (blue), and~Medium-High (green). \textbf{(a)} Displays the cluster distribution considering PGP, ERPM and GP. \textbf{(b)} Considers PGP, ERPM and XACC var. \textbf{(c)} Displays clusters regarding PGP, GP and XACC var, and \textbf{(d)} considers ERPM, GP and XACC var. }
    \label{fig:3D_3clusters}
\end{figure}

First, we carried out a~three-cluster grouping of the SOM neurons. Table~\ref{tab:drivers_3clusters} presents relevant statistical values of fuel consumption for each cluster: average value, variance, and~maximum value. Taking these values into account, the~clusters were labeled as Very Low, Low, and Medium-High. The~classification results applied to the Uyanik dataset are shown in Figure~\ref{fig:3D_3clusters}, where three-dimensional views are~provided.

\begin{table}[H]
\caption{Percentage of the route that each driver travels using different fuel-consumption driving styles (DSs) (three-cluster classification).}
\label{tab:drivers_3clusters}
\centering
\begin{tabular}{lccc}
\toprule
\textbf{Driver}	& \textbf{Very Low (\%)} & \textbf{Low (\%)}& \textbf{Medium-High (\%)}\\
\midrule
{\hl{D1}}  &  4.7   &  20.0  &  \textbf{75.3}\\
{\hl{D2}}  &  49.4  &  10.4  &  40.2\\
{\hl{D3}}  &  23.0  &  33.3  &  43.7\\
{D4}  &  7.8   &  51.1  &  41.1\\
{D5}  &  33.3  &  26.5  &  40.2\\
{D6}  &  \textbf{80.2}  &  10.4  &   9.4\\
{D7}  &  15.1  &  33.7  &  51.2\\
{D8}  &  16.2  &  29.7  &  54.1\\
{D9}  &  8.1   &  38.4  &  53.5\\
{D10} &  56.3  &      0  &  43.7\\
{D11} &  \textbf{78.6}  &  11.9  &   9.5\\
{D12} &  10.6  &  33.0  &  56.4\\
{D13} &  14.0  &  46.5  &  39.5\\
{D14} &   8.4  &  25.3  &  \textbf{66.3}\\
{D15} &   2.5  &  45.7  &  51.8\\
{D16} &  14.3  &  40.5  &  45.2\\
{D17} &  38.1  &  21.4  &  40.5\\
{D18} &   2.7  &  48.0  &  49.3\\
{D19} &  15.6  &  30.0  &  54.4\\
{D20} &  32.2  &  26.9  &  40.9\\
\bottomrule
\end{tabular}
\end{table}
%MDPI: Is the bold necessary? If yes, please add explain for bold. Please confirm and check below.
%Author: We removed the bold for the first column. On the other hand, the numbers in bold may be helpful for readers to better understand the table.

The~three displayed clusters are compact, their contained data are contiguous, and~they are clearly separated, as can be seen in Figure~\ref{fig:3D_3clusters}. Matching clusters with their associated consumption displayed in Table~\ref{tab:consumption_3clusters} by color, it is apparent that the green cluster corresponds to medium-high fuel consumption rides, the~blue one to low consumption, and~the red group represents very low fuel consumption rides. Additionally, by~analyzing clusters' fuel consumption variances, it can be seen that the higher the average value, the~higher the variance, with this correlation being a~noticeable feature of the identified~groups.

Further analysis of the relationships of the identified groups with the driving features displayed in each sub-figure can be performed. Regarding Figure~\ref{fig:3D_3clusters}a,b, the~DS groups look similar, since the GP variable of (a) and the XACC var of (b) are highly correlated as a~measure of swift operation of the gas pedal. On~the other hand, Figure~\ref{fig:3D_3clusters}c shows a~different cluster distribution. In~this graph, the~Low and Very low consumption classes (blue and red, respectively) are interleaved. This happens because the correlation between XACC var and GP is strong, with~GP vs. XACC var providing no additional meaningful information. In~contrast, the~PGP vs. GP and the PGP vs. XACC var planes show that the positioning of the clusters is interchanged with respect to Figure~\ref{fig:3D_3clusters}a,b. Nevertheless, this interchange is coherent with the precedent figures, since the green cluster is placed at the upper range of the PGP axis, while the other ones are at the lower range, the~blue cluster is related to low GP, and the red one is related to medium GP. The~same clusters' position interchange phenomenon can be observed in Figure \ref{fig:3D_3clusters}d, according to the aforementioned characteristics.

When considering the cluster distribution of Figure~\ref{fig:3D_3clusters}, and~taking into account that aggressiveness and fuel consumption are well correlated, considering this figure as our baseline for further comparisons, we can assert~that
\begin{itemize}[leftmargin=*,labelsep=5.8mm]
    \item Very low fuel consumption (red) corresponds to drivers who keep the car running at its lowest regime  (low PGP, low ERPM, and~medium GP).
        \item Low fuel consumption (blue) corresponds to drivers who use the gas pedal gently and run the car at medium regimes  (low PGP, low GP, and~medium ERPM).
    \item Medium-High fuel consumption (green) corresponds to drivers who use the gas pedal extensively and run the car at high engine regimes  (high PGP, disperse GP, and~high ERPM).

\end{itemize}

\begin{table}[H]
\caption{Fuel consumption (L/100km) parameters of the three-cluster~classification.}
\label{tab:consumption_3clusters}
\centering
\begin{tabular}{lccc}
\toprule
\textbf{Cluster Label}	& \textbf{Average Value} & \textbf{Variance}& \textbf{Maximum Value}\\
\midrule
{\hl{Very low (red)}}	& 2.76 & 1.02 & 6.66\\
{\hl{Low (blue)}} & 3.04 & 1.53 & 7.80\\
{\hl{Medium-High (green)}} & 5.15& 3.34 & 12.6\\
\bottomrule
\end{tabular}
\end{table}
\unskip

%MDPI: Please add explanation for (a)--(d). Please confirm and check below.
%Author: Please check that we added the required explanation.

Nevertheless, despite interesting DS-related fuel consumption profiles being extracted at a~joint interpretation of the information that is depicted in Figure~\ref{fig:3D_3clusters} and Table~\ref{tab:consumption_3clusters}, driver classification cannot be kept uniform along an~entire trip, since it is far from being a~binary task. For~that reason, due to driving circumstances changing during a~trip, evaluation by time windows provides a~better assessment of the fuel consumption trend. In~Table~\ref{tab:drivers_3clusters}, the~distribution of DSs among clusters is displayed. As~can be seen, each driver shows a~unique cluster distribution for his/her trip. This distribution means that fuel-consumption-related DS is not a~binary feature, but~a composition of several cluster mixture~ratios.

Four drivers stand out among the remaining ones: D1, D6, D11, and~D14, as remarked in bold in Table~\ref{tab:drivers_3clusters}. Thus, D6 and D11 spend a~longer time classified with Very low consumption DS, with $80.2\text{\%}$ and the 78.6\% of the total ride time, respectively, so they can be considered as eco-friendly drivers. On~the other hand, D1 and D14 are the opposite case, with 75.3\% and 66.3\% of the total ride time being classified as Medium-High fuel consumption drivers, totally compromising eco-friendliness. According~to the clusters identified in Figure~\ref{fig:3D_3clusters}, while the former drivers operate the throttle pedal uniformly and keep engine RPMs low, being an~ideal operation decision, the~latter ones swiftly operate the gas pedal and keep engine RPMs at the upper range for most of the trip. Finally, it is worth remarking that most drivers' behavior evolves between contiguous classes, except~D10, which exhibits a~particular behavior, leaping between extreme classes (from Very low to Medium-High, and vice~versa).

Table~\ref{tab:clusters_action} compiles the actions drivers should perform to modify their DS with the aim of reducing their fuel consumption attending to their current classification. Different~actions are needed, depending on the group, as~can be seen in this table. Thus, for~example, since Medium-High fuel consumers typically operate the gas pedal swiftly, keeping RPMs high due to that aggressiveness, they are required to lower RPMs while trying to operate the gas pedal smoothly and to a~lesser extent. On~the other hand, low consumption drivers are required to switch to a~higher gear because, despite~their softly operating the gas pedal, they keep RPMs high due to the use of low gears. Finally,~very~low fuel consumers are required to keep their DSs with no~changes.

\begin{table}[H]
\caption{Actions that are associated to the three-cluster~classification.}
\label{tab:clusters_action}
\centering
\begin{tabular}{lc}
\toprule
\textbf{Current Cluster}	& \textbf{Required Action}\\
\midrule
{\hl{Very low (red)}}	& Keep driving style\\
{\hl{Low (blue)}} & Lower RPM/Switch to a~higher gear\\
{Medium-High (green)} & Lower RPM/Keep gas steady/Lower PGP\\
\bottomrule
\end{tabular}
\end{table}

\paragraph{Five-Cluster Grouping}

\begin{figure}[H]
    \centering
    \includegraphics[width=0.70\textwidth]{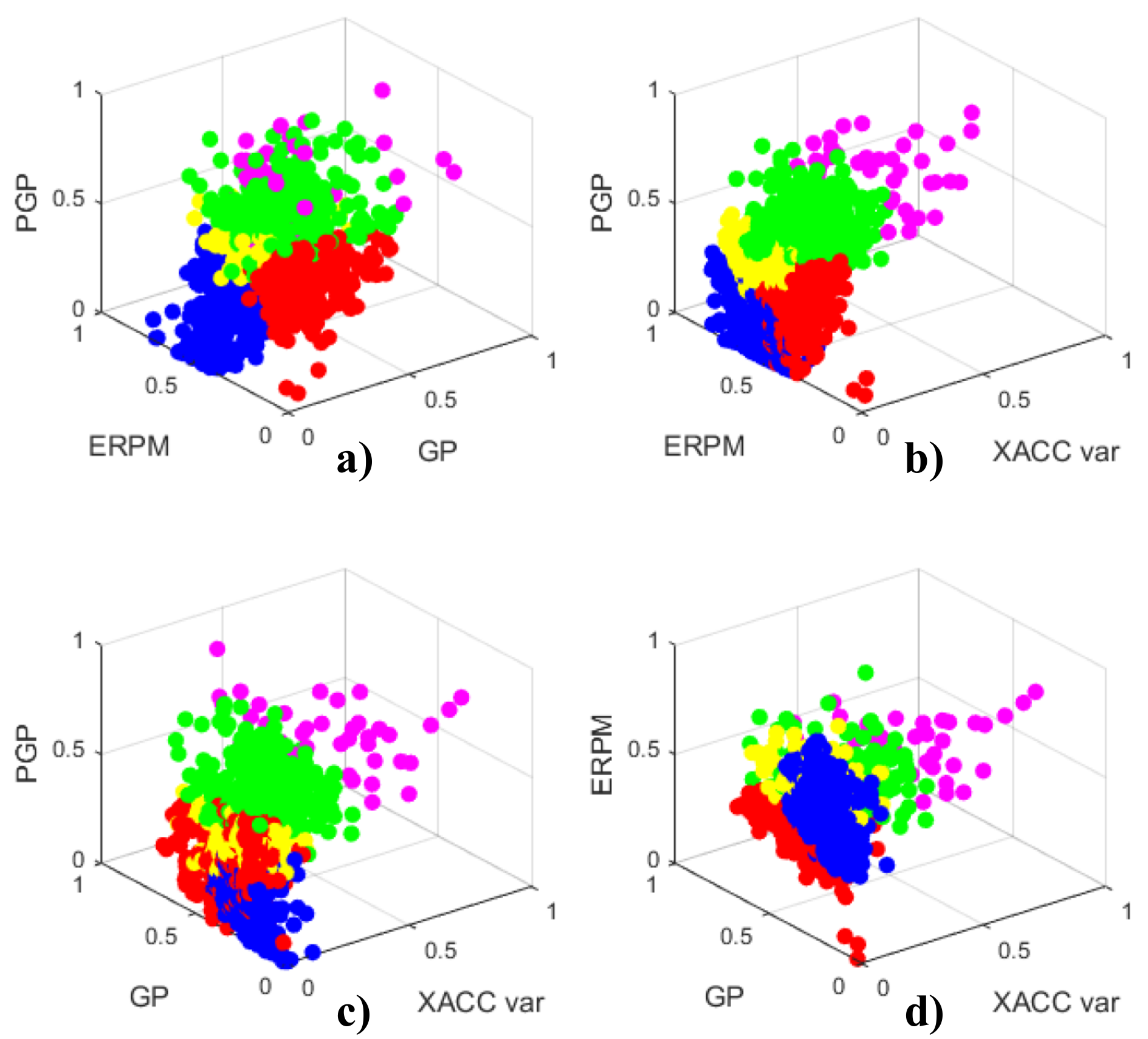}
    \caption{\hl{Three-dimensional }views of the five-cluster fuel consumption classification results. The~clusters were labeled as Very low (red), Low (blue), Medium (yellow), High (green), and~Very~high~(magenta).  \textbf{(a)} Displays the cluster distribution considering PGP, ERPM and GP. \textbf{(b)} Considers PGP, ERPM and XACC var. \textbf{(c)} Displays clusters regarding PGP, GP and XACC var, and \textbf{(d)} considers ERPM, GP and XACC var.}
    \label{fig:3D_5clusters}
\end{figure}

The recommendations that are indicated in Table~\ref{tab:clusters_action} could be unclear for some drivers, especially those being classified into the green cluster (Medium-High consumption). For~that reason, with~the aim of personalizing the driving recommendations, SOM clustering of the trained map was recomputed using a~lower threshold value in its U-matrix, so that more precise partitions could be obtained. Figure~\ref{fig:3D_5clusters} and Table~\ref{tab:consumption_5clusters} show the five-cluster grouping obtained after this re-computation and the associated fuel consumption for each cluster, respectively. Nevertheless, despite the existence of a~higher number of groups, the~relationships identified in Figure~\ref{fig:3D_3clusters} remain. Thus, the~blue and red clusters (Low and Very low consumption) are kept barely unaltered %this means altered so much that it is barely the same, please confirm
both in position and number of elements. In~contrast, three new classes appear from the former Medium-High consumption group, namely Medium, High, and~Very High (yellow, green, and~magenta, respectively).

\begin{table}[H]
\caption{\hl{Fuel consumption} (L/100km) parameters of the five-cluster~classification.}
\label{tab:consumption_5clusters}
\centering
\begin{tabular}{lccc}
\toprule
\textbf{Cluster Label}	& \textbf{Average Value} & \textbf{Variance}& \textbf{Maximum Value}\\
\midrule
{\hl{Very low (red)}}	& 2.75 & 1.04 & 6.66\\
{\hl{Low (blue)}}& 3.04 & 1.54 & 7.80\\
{\hl{Medium (yellow)}}& 4.44 & 2.21 & 10.1\\
{High (green)}& 5.42 & 3.13 & 11.4\\
{Very high (magenta)}& 7.81 & 5.38 & 12.5\\
\bottomrule
\end{tabular}
\end{table}
\unskip
%MDPI: Please add explanation for (a)--(d).

Being the aforementioned partitions, extracted from Figure~\ref{fig:3D_5clusters}, jointly analyzed with the statistical data contained in Table~\ref{tab:drivers_5clusters}, we can assert that this SOM-based grouping is finer, and~more meaningful information can be extracted when compared to the three-cluster classification. 

\begin{table}[H]
\caption{Percentage of the route that each driver travels using different fuel-consumption DS (five-cluster classification).}
\label{tab:drivers_5clusters}
\centering
\begin{tabular}{lccccc}
\toprule
\textbf{Driver}	& \textbf{Very Low (\%)} & \textbf{Low (\%)}& \textbf{Medium (\%)} & \textbf{High (\%)} & \textbf{Very High (\%)}\\
\midrule
{\hl{D1}}  &  4.7   &  20.0  &  \textbf{\hl{44.7}}  &  \textbf{\hl{30.6}}  &  0\\
{\hl{D2}}  &  48.3  &  10.3  &  2.3   &  32.2  &  6.9\\
{D3}  &  21.8  &  33.3  &  13.8  &  30.0  &  1.1\\
{D4}  &  7.8   &  51.1  &  22.2  &  18.9  &   0\\
{D5}  &  33.3  &  26.5  &  10.3  &  25.3  &  4.6\\
{D6}  &  \textbf{71.9}  &  10.4  &   9.4  &  8.3   &  0\\
{D7}  &  14.0  &  33.7  &  18.6 &  29.1   &  4.6\\
{D8}  &  14.9  &  29.7  &   8.1 &  37.8   &  9.5\\
{D9}  &  8.1   &  38.4  &  31.4 &  22.1   &  0\\
{D10} &  56.3  &     0  &  16.1 &  25.3   &  2.3\\
{D11} &  \textbf{78.6}  &  11.9  &   1.2 &  8.3    &  0\\
{D12} &  10.6  &  33.0  &  37.2 &  19.2   &  0\\
{D13} &  14.0  &  46.5  &  15.1 &  24.4   &  0\\
{D14} &   8.4  &  25.3  &  \textbf{31.3} &  \textbf{35.0}   &  0\\
{D15} &   2.5  &  45.7  &  24.7 &  23.4   &  3.7\\
{D16} &  13.1  &  40.5  &  30.9 &  15.5   &  0\\
{D17} &  38.1  &  21.4  &  10.7 &  21.5   &  8.3\\
{D18} &   2.7  &  48.0  &  19.2 &  26.0   &  4.1\\
{D19} &  15.6  &  30.00 &  21.1 &  32.2   &  1.1\\
{D20} &  32.2  &  26.9  &  22.6 &  18.3   &  0\\
\bottomrule
\end{tabular}
\end{table}
\unskip
%MDPI: Is the bold necessary? If yes, please add explain for bold. 
%Author: We removed the bold for the first column. On the other hand, the numbers in bold may be helpful for readers to better understand the table.

%MDPI: Please insert it close to where it is first mentioned in the text. 
%Author: We re-distributed the paragraphs to fulfill the request.

The~fuel-consumption-associated DS ratios for each driver are recalculated in Table~\ref{tab:drivers_5clusters} in order to~verify that the five-cluster classification adds information to the already existing groupings, mainly with the purpose of personalizing the assessment of the most aggressive drivers. In~this case, D1, D6, D11, and~D14 are analyzed again to inspect whether the new cluster classification changes the information contained in Table~\ref{tab:drivers_3clusters}. For~these drivers, the~percentages of Very low and Low consumption remain practically unchanged with respect to the three-cluster table. On~the other hand, if~we accumulate the percentages of medium, high, and~very high consumption instants, we can observe that they practically match the Medium-High column of Table~\ref{tab:drivers_3clusters}. This means that not only does this five-cluster classification provide comparable results, but~it also allows one to thoroughly examine the detailed behavior formerly grouped as medium-high consumption~DS.

According to Figure~\ref{fig:3D_5clusters}, the~Medium-High consumption cluster of Figure~\ref{fig:3D_3clusters} can be detailed with the following groups:

\begin{itemize}[leftmargin=*,labelsep=5.8mm]
    \item Medium fuel consumption (yellow): corresponds to drivers who run the car at engine regimes similar to those achieved for the low consumption cars, but~with the difference of a~more extensive use of gas pedal, (i.e., medium PGP, low GP and medium ERPM).
    \item High fuel consumption (green): corresponds to drivers who run the car at medium-high RPM, with~moderate swiftness of the gas pedal operation (medium-high ERPM, medium-high PGP, and medium GP).
    \item Very high fuel consumption (magenta): corresponds to drivers who are slightly more aggressive that those from the preceding group (high ERPM, high PGP, and medium-high GP).

\end{itemize}

Additionally, with~this new cluster distribution, more actions can be indicated to drivers to modify their DS. Thus, in~contrast with Table~\ref{tab:clusters_action}, where Medium (yellow) to Very high (magenta) consumption classes were aggregated, in~\hl{Table}~\ref{tab:5clusters_action}, actions were added for each individual newly identified cluster. In~contrast, actions for Very low and Low consumption groups remain unchanged. The~action Lower RPM/Keep gas steady was transformed into the sequence Lower RPM/Operate gas softly $\rightarrow$ Lower PGP/Lower RPM $\rightarrow$ Lower PGP/Keep gas steady, as~can be noticed when comparing Table~\ref{tab:clusters_action} with Table~\ref{tab:5clusters_action}. This happens because, differing from the uniform DS of the big Medium-High consumption group of the three-cluster classification (green group in Figure~\ref{fig:3D_3clusters}), the~Very high (magenta) fuel-consumers are required to lower both RPM and PGP, because they drive at high RPM. In~contrast, High (green) consumers run engines at moderated RPM rates. However, the~latter swiftly operate the gas pedal at moderately high percentages, consequently being required to use it less and more smoothly. Finally,~Medium (yellow) consumers operate the gas pedal smoothly but their usage percentage is still high, so they are required to further reduce gas pedal~usage.
%MDPI: Table citation should be in numerical order, Table 8 is cited before Table 7, please reconfirm.
%Author: Please check that we moved Table 6 onto Figure 10 to provide a structure coherent with Table 4+Figure 9. Additionally, 5 lines below Table 5 we modified the sentence to mention Table 6 and to provide more meaning to the re-distribution observed in Figure 10.

\begin{table}[H]
\caption{Actions associated to the five-cluster~classification.}
\label{tab:5clusters_action}
\centering
\begin{tabular}{lc}
\toprule
\textbf{Current Cluster}	& \textbf{Required Action}\\
\midrule
{\hl{Very low (red)}}	& Keep driving style\\
{\hl{Low (blue)}} & Lower RPM/Switch to a~higher gear\\
{\hl{Medium (yellow)}} & Lower RPM/Operate gas softly\\
{High (green)} & Lower PGP/Lower RPM\\
{Very high (magenta)} & Lower PGP/Keep gas steady\\
\bottomrule
\end{tabular}
\end{table}

Should drivers follow the recommendations that are displayed in Table~\ref{tab:5clusters_action}, a~noticeable reduction in fuel consumption is expected to occur. However, because the level of engagement of motorists with the provided advice may vary depending on behavioral characteristics of each individual, the~expected improvement on fuel economy must be cautiously analyzed. For~that reason, Table~\ref{tab:perc_reduction} was elaborated to estimate the expected improvement when considering a~minimal level of engagement with the system that would allow drivers to modify their DS to the immediately adjacent~cluster. 

\begin{table}[H]
\caption{Expected fuel-consumption reduction between contiguous~clusters.}
\label{tab:perc_reduction}
\centering
\begin{tabular}{llc}
\toprule
\textbf{Current Cluster} & \textbf{Target Cluster} & \textbf{Reduction (\%)}  \\ \midrule
{\hl{Low}}             & {\hl{Very low}}       & 9.5               \\
{\hl{Medium}}          & {Low}            & 31.5              \\
{High}            & {Medium}         & 18.1             \\ \bottomrule
\end{tabular}
\end{table}
 
As can be seen in Table~\ref{tab:perc_reduction}, obtained from the values displayed in \hl{Table}~\ref{tab:consumption_5clusters}, if~drivers could only improve their DS to the best adjacent class, reductions in fuel consumption ranging from 9.54\% up 31.5\% are expected, with~even higher performances for strongly-engaged drivers, showing that a~significant reduction in polluting agents could be expected if this system was implemented in cars. This~potential reduction is significantly greater than the already existing systems, which were exposed in Section~\ref{sec:Related_res}, making this approach a~promising~solution.

\section{Fuel-Consumption Assessment~Results}
\label{sec:results}

 Most drivers' behaviors vary among different clusters, and,  the~Very low consumption class being the ideal one, indications should be addressed to drivers to modify their DS if they fall into the other classes at any moment of the ongoing ride, as has been seen in the previous section. In~the following, the~driving behavior of two particular drivers, D1 and D11, will be analyzed with the aim of verifying the suitability of the advice provided by the fuel-consumption assessment system. In~addition, the~potential impact on emissions that this advice system can achieve is quantitatively~evaluated. 

\subsection{Drivers'~Advice}

\begin{figure}[H]
    \centering
    \includegraphics[width=0.9\textwidth]{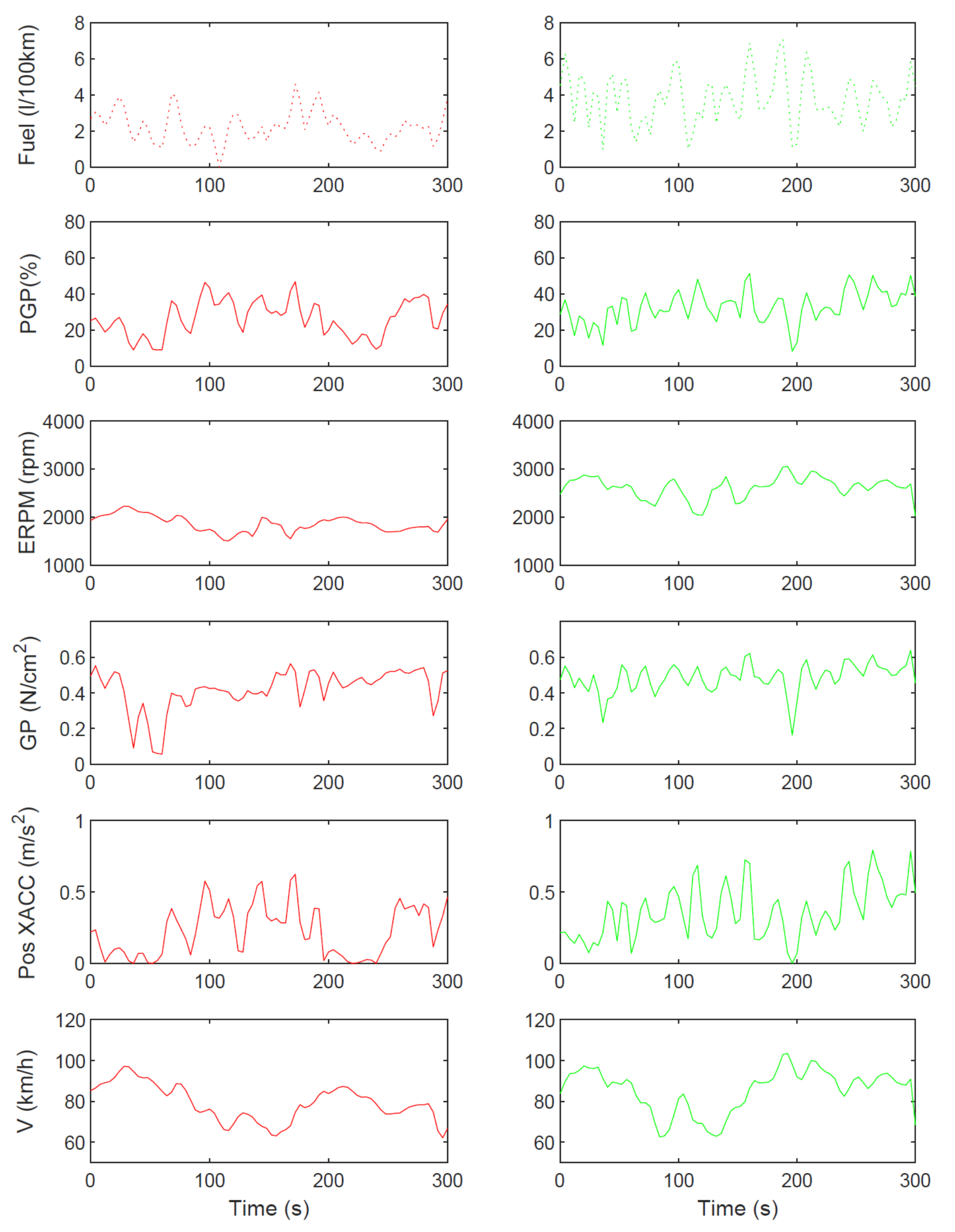}
    \caption{\hl{Uyanik measurement} of relevant CAN-bus and IMU signals corresponding to  D1 (green) and D11 (red) during five uninterrupted minutes of the route. The~driving behavior was evaluated every 100 s, and~the cluster with the maximum percentage was selected. Both D11 and D1 were classified into a~single cluster during the whole segment of the trip: D11 drives according to the  Very low cluster, and~D1's driving style is mostly into the Medium cluster. GT-Suite simulations of fuel consumption are also~displayed.}
    \label{fig:time_series}
\end{figure}

\begin{figure}[H]
    \centering
    \includegraphics[width=0.99\textwidth]{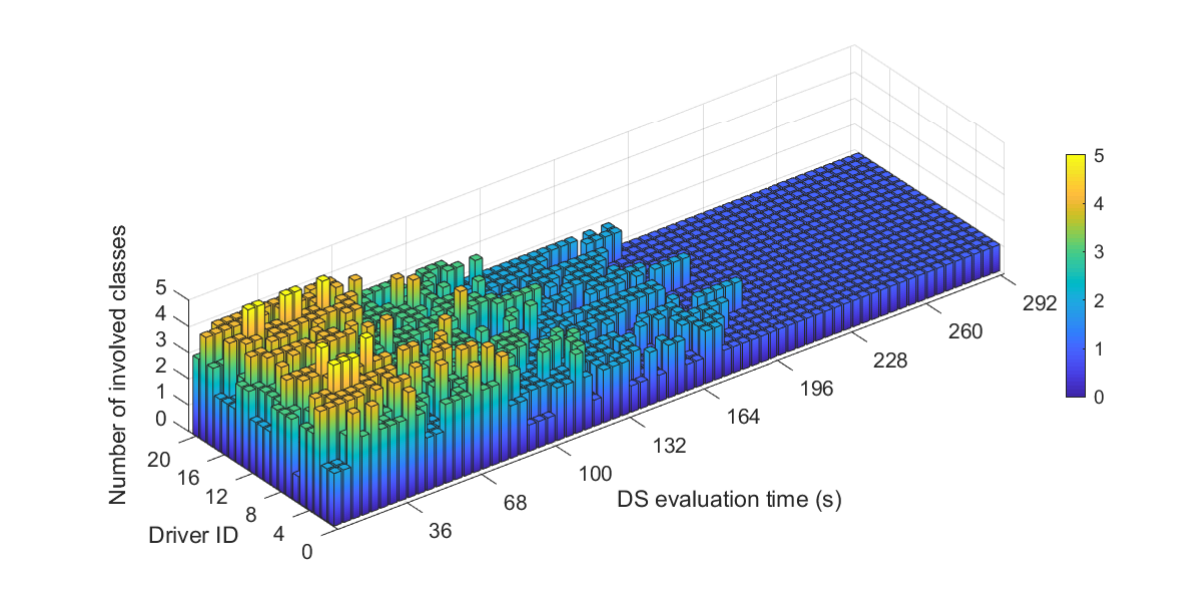}
    \caption{Three-dimensional bar diagram of the number of classes identified, depending on the evaluation time for each~driver.}
    \label{fig:histo}
\end{figure}

Figure~\ref{fig:time_series} depicts the Uyanik measurement of relevant CAN-bus and IMU signals corresponding to D1 (green) and D11 (red) during five uninterrupted minutes of the route. DS evaluation times from 8 s to 292 s were considered for the performance analysis, as~can be seen in Figure~\ref{fig:histo}. It can be observed that the longer the evaluation time, the~lower the number of involved DS classes for each evaluation period. For~many drivers, this number of classes reached a~local minimum at a~length of 100 s, while overshooting for longer times until the low-pass filtering effects of a~very long evaluation time happened, which would eliminate the details that are needed for a~correct DS evaluation. For~that reason and, according to~\cite{Azadani2020}, the~corresponding driving behavior was evaluated every 100 s in order to be correctly assessed and to provide useful advice to the drivers, with~the last 100 s stretch being test data (i.e.,~unseen by the system). Taking into account the evolution of the DS during each 100 s stretch, the~cluster with the maximum percentage was selected. D11 and D1 were both classified into a~single cluster during the whole segment of the trip: D11 drove according to  the   Very low cluster, and~D1's driving style was mostly in the Medium cluster. GT-Suite simulations of fuel consumption are also displayed. As~can be seen, the~average fuel consumption of D11 is much lower than D1, as~expected. Again, the~measured RPMs are lower for D11 than for D1, and~the same is the case for the variance of the positive XACC, as~in the cluster distribution of Figure~\ref{fig:3D_5clusters}. On~the other hand, measured speeds are not significantly different, proving that fuel consumption under similar conditions has more to do with the car handling itself than with speed. In~consequence, the~eco-driving system would provide the following advice: D11: ``Keep driving style''; D1: ``Lower RPM/Operate gas~softly''.

\unskip
%MDPI: Please confirm if it is necessary to add a, b,...,l in the sub-figures and add explanation for them in the caption.
%Author: In this case it's not necessary because each column corresponds to several variables for a given driver (red for a driver and green for the other one)

In sum, most drivers' classifiable behaviors vary among different clusters and, with~the Very low consumption class being the ideal one, indications should be addressed to drivers to modify their DS if they fall into the other four classes at any moment of the ongoing ride. The~eco-driving system provides advice to the driver according to a~user-configurable time~interval.

\subsection{Fuel Consumption and Emissions~Reduction}

The~same two drivers (D1 and D11) shown in Figure~\ref{fig:time_series} were selected in order to indicate the potential reduction on emissions that this advice system can achieve. In~this 300 s highway driving stretch, speeds~for both drivers are kept above 79 %please check
 km/h (79.1 km/h and 85.4 km/h, respectively). In~these conditions, the~DS identification system classifies D1 mostly into the Medium consumption class, while~D11 is classified into the Very low consumption class. Additionally, the GT-Suite simulation data show that the mean fuel consumption measurements in that stretch for D1 and D11 are 4.46~\hl{L}/100 km and 2.61 \hl{L}/100 km, respectively. Assuming that the average composition for diesel fuel corresponds to the formula $\text{C}_{12}\text{H}_{23}$, with~a density of 0.835 g/\hl{L}~\cite{Prasanna2019}, the~stoichiometric combustion of this fuel type follows the \hl{equation}
%MDPI: Please check if this should be a L.
%Author: Yes, it should be L, because we were talking about liters, thank you for detecting this typo.
%MDPI: Please confirm is the italics in Euqation (7) necessary.
%Author: It was not necessary, so, according to your instructions, we removed it, thank you

\begin{equation}
  {4 \text{C}}_{12}\text{H}_{23}{ + 71 \text{O}}_{2}{ \rightarrow 48 \text{CO}}_{2} +{ 46 \text{H}}_{2}\text{O.}
    \label{eq:comb}
\end{equation}

With the chemical reaction of Equation \eqref{eq:comb}, the~average CO$_2$ generation rate can be calculated, with~the CO$_2$ emissions for D1 and D11 being 128.4 g/km and 75.2 g/km, respectively. That is to say,   D11's CO$_2$ emissions are 41.4\% lower than those of D1 for similar road stretches at similar speeds. With~these results, we can assert that, if~the recommendations of Tables~\ref{tab:clusters_action} and \ref{tab:5clusters_action} were provided to D1, with~the aim of being classified into the Very low cluster, then a~noticeable reduction in fuel consumption and emission rates could~happen.

In sum, by~using SOM algorithms, the~clusters of Figures~\ref{fig:3D_3clusters} and \ref{fig:3D_5clusters} were discovered and fuel-consumption-associated DS features were extracted. Those clusters and features, jointly with the analysis of the cluster distribution ratio for each driver of Tables~\ref{tab:drivers_3clusters} and \ref{tab:drivers_5clusters}, allowed for the identification of complex behaviors. Finally, several actions were described to modify individual DSs with the aim of improving fuel economy while taking the clustering as well as the distributions and the features into account, consequently encouraging~eco-driving.

\section{Implementation of the PSoC-Based Intelligent~System}

\label{sec:implementation}

The intelligent system for real-time assessment of fuel consumption and eco-driving was properly evaluated and tested through a specific PC-based model. After~that, the~whole system was implemented, such that it can be executed in real-time. For~that purpose, the~device that this task is implemented in must be capable of performing high-speed data computing, while providing high throughput data outputs. For~those reasons, a~hybrid HW/SW architecture was developed and implemented on the Xilinx XC7Z045-2FFG900 Programmable PSoC~\cite{DS190} using the Xilinx ZC706 development board~\cite{UG954}.  

\begin{figure}[H]
    \centering
    \includegraphics[width=0.6\textwidth]{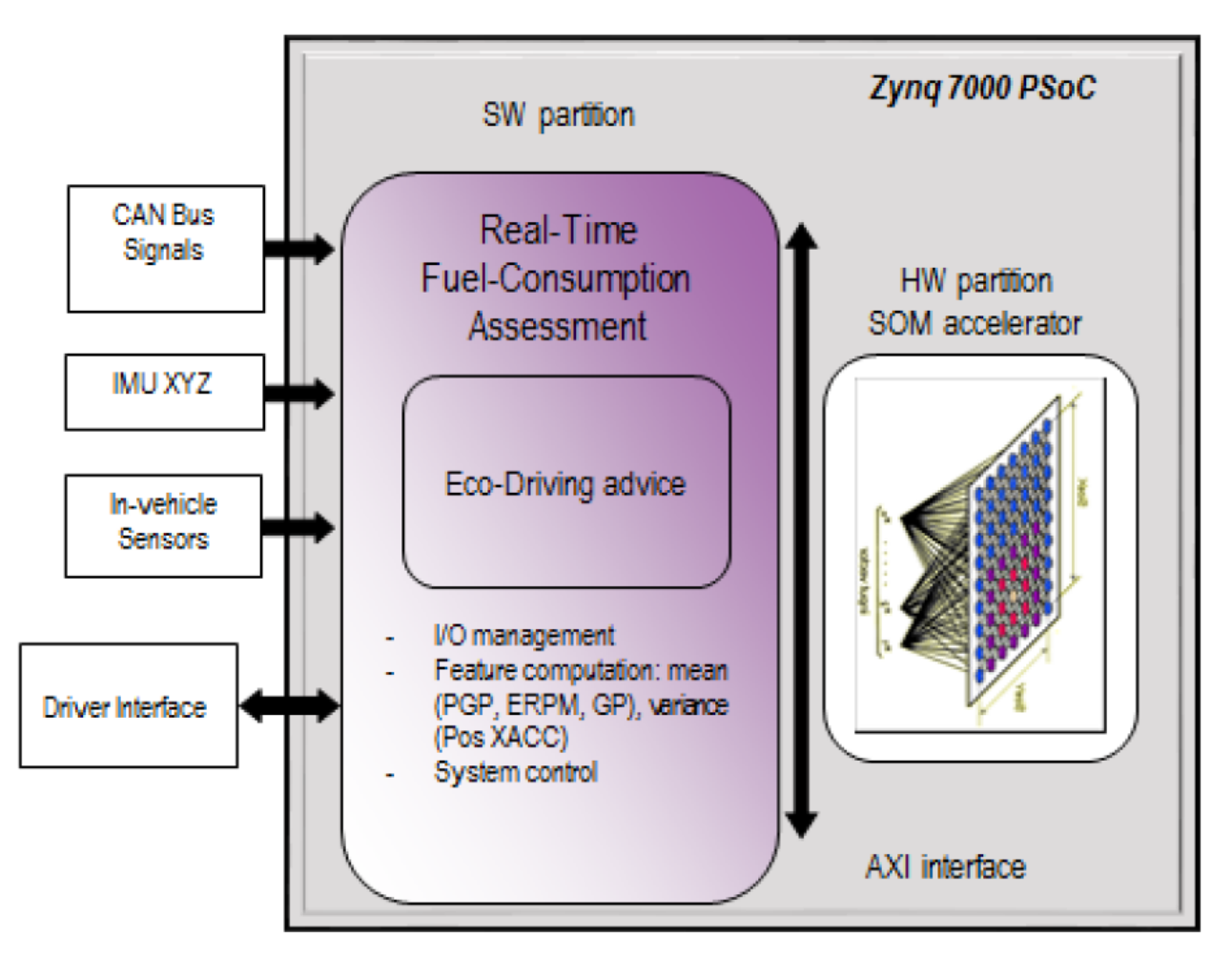}
    \caption{Block diagram of the programmable system-on-a-chip (PSoC) for real-time fuel consumption assessment and~eco-driving. }
    \label{fig:psoc_blocks}
\end{figure}

Figure~\ref{fig:psoc_blocks} depicts a~block diagram of the proposed solution. PSoCs are devices that combine the high performance of FPGAs with the flexibility of embedded microprocessor-based systems that are both connected with each other by means of an~internal AXI4 interface. Thus, while the FPGA (HW~partition) performs highly parallelizable binary operations, the~microprocessor (SW partition) is specially dedicated to executing sequential tasks, being enabled to interchange information through the AXI4~bus.

The entire HW partition of the system, consisting of an~SOM accelerator, was deployed in the FPGA of the PSoC using VHDL language and the Xilinx Vivado 2018.1 design suite~\cite{Vivado2018}. On~the other hand, the~remainder of the proposed system functionalities were programmed at the microprocessor (SW partition depicted in Figure~\ref{fig:psoc_blocks}) by developing a~bare-metal C application that can acquire data from the buses of the vehicle, compute the windows of the ERPM, GP, PGP, and~pos XACC features, share them with the FPGA, retrieve the SOM accelerator results, and provide advice to drivers. %please~check

\subsection{Hardware Partition: SOM~Accelerator}
\label{subsec:HW}
The hardware partition's top-bottom hierarchy, deployed within the PSoC's FPGA, can be described, as~follows.

\begin{figure}[H]
    \centering
    \includegraphics[width=0.75\textwidth]{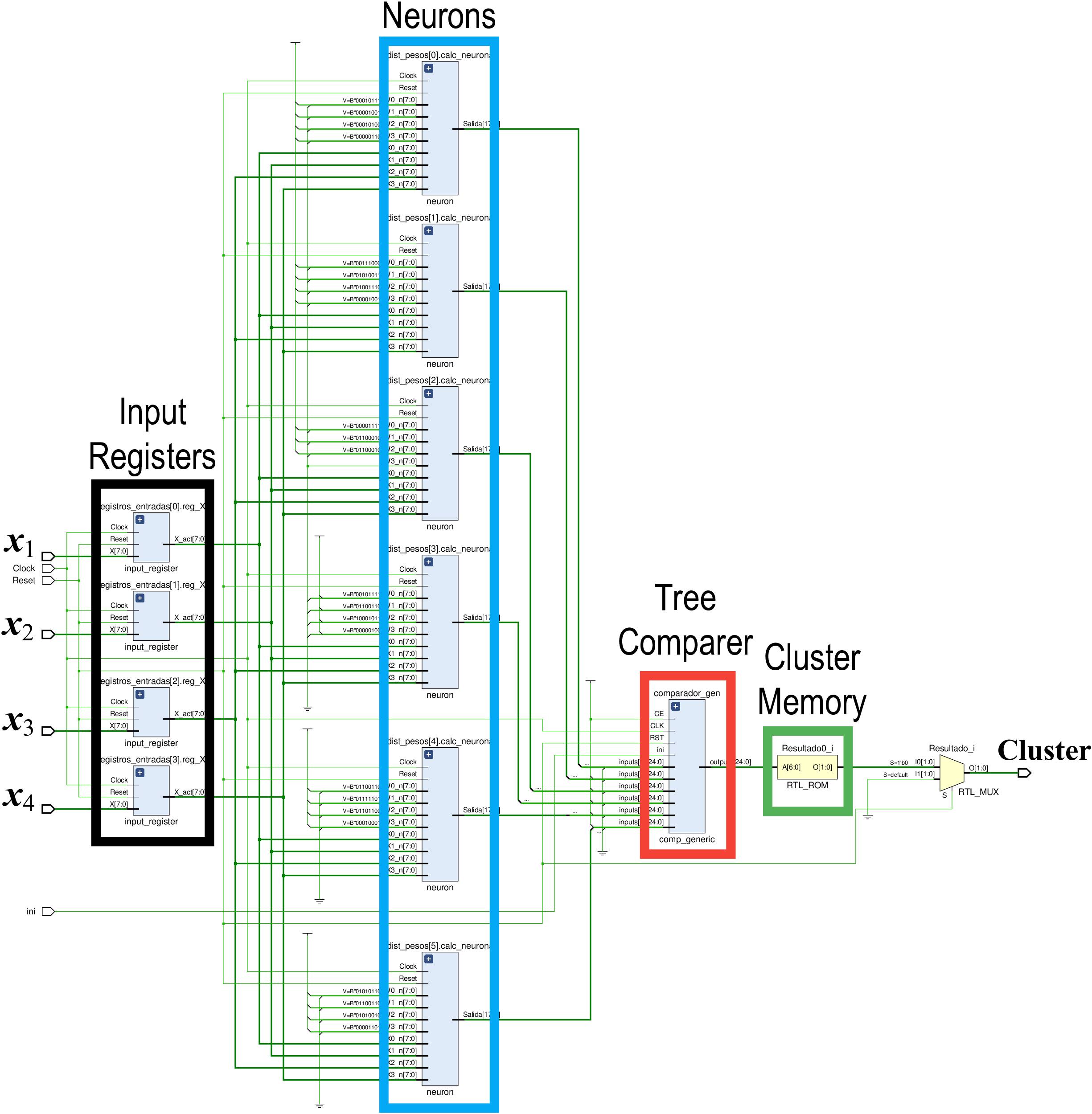}
    \caption{\hl{Scheme of} the SOM hardware accelerator. A~four-input SOM topology with six output neurons is~shown as~a case example.}
    \label{fig:arq_global}
\end{figure}

The SOM hardware accelerator is composed of four main modules: input registers, neurons, comparers, and~internal ROMs (see Figure~\ref{fig:arq_global}), as~well as a~controller unit. This architecture has been designed to be totally parallel with the aim of returning a~correct response in the minimum time lapse. The~VHDL language is used in order to create a~fully scalable architecture regarding the number of input features ($N$) and the number of neurons of the SOM ($M$).

\unskip
%MDPI: Please replace with a sharper image.
%Author: We converted the image to PNG to increase sharpness at low zoom levels.

\subsubsection{Input~Registers}
The input registers (the black box in Figure~\ref{fig:arq_global}) are used in order to feed the input samples \mbox{$\mathbf{x}=\left(x_1,x_2,\dots,x_N\right)$} into the SOM accelerator synchronously, with~each rising edge of the clock signal. The~number of input registers depends on the number of input features, $N$, since each feature needs a~separate register. These registers' inputs are read from the AXI4~interface.

\subsubsection{Neurons}

The neuron components (the blue box in Figure~\ref{fig:arq_global}) compute the squared Euclidean distance between the input sample and a~given neuron's weight (see Equation \eqref{eq:Euclidean}). Each neuron block is shaped by two types of components: the distance module and the adder module. Thus, while the former computes how far each input feature $x_j^k$ is from the corresponding neuron weight $m_{ij}$ and squares that difference (squared euclidian distance), the~latter, which is based on a~typical tree-adder, sums the $N$ individual squared distances to compute the total distance from the input sample to the $i$-th neuron~weights.

Besides, a~neuron pointer, $i$, is added to the output. It indicates which neuron each computed distance belongs to, with the aim of easily accessing the cluster memory once the neuron with the minimum distance (i.e.,~the BMU) is found. Finally, each neuron block stores its corresponding weights into a~small~ROM.

\subsubsection{Tree-Comparer}

\begin{comment}
For this reason, comparers are used to compute, which is the neuron holding the shortest distance to the input sample. These modules compare the neurons' outputs in %please check
  couples until they find the one with the shortest~distance. 

The neurons' weights as well as the clusters to which they belong, once the SOM training has been completed and the neurons are correctly classified, are stored into an~internal ROM placed within the FPGA architecture. Besides~this, a~reset signal is used to erase all the registers, and~an enable signal  %please check
 allows the operation of the controller.
\end{comment}

\begin{figure}[H]
    \centering
    \includegraphics[width=0.55\textwidth]{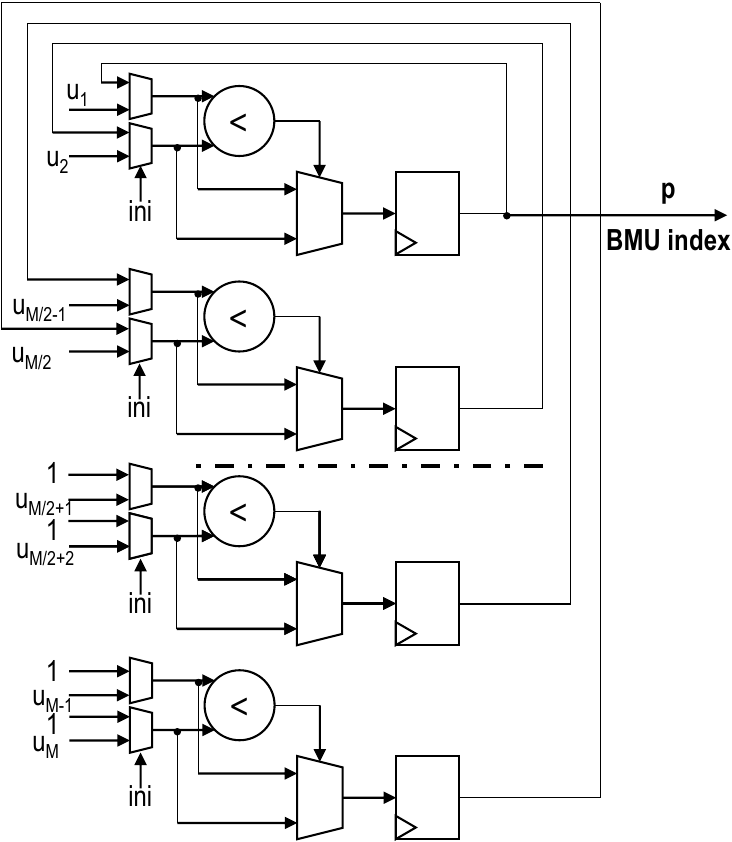}
    \caption{Scheme of the proposed recursive tree comparer architecture that substitutes a~traditional comparer~solution.}
    \label{fig:comp_struc}
\end{figure}

The tree comparer (the red box in Figure~\ref{fig:arq_global}) computes the BMU of the input sample (see~Equation~\eqref{eq:BMU}). The~input to the module is the array of outputs of the neuron blocks, that is to say, the~distances between each neuron with the input sample, concatenated with the neuron pointers. It returns the BMU along with its corresponding pointer (i.e.,~the BMU index). For~this design, a~recursive tree-comparer was developed based on the previous work of the authors~\cite{mata2019hardware}. This~topology, as shown in Figure~\ref{fig:comp_struc}, was adapted so as to provide latency rates that were comparable to those from traditional binary tree comparers, while minimizing resource~usage. 

Thus, the~proposed architecture of Figure~\ref{fig:comp_struc} only uses $M/2$ comparer blocks, while a~binary tree comparer would spend $M-1$ of the same hardware resources. Given an~input array \mbox{$\mathbf{u}=\left(u_1,u_2,\dotsc,u_M \right)$}, and~a control control signal \texttt{ini}, the~module operates, as follows:

\begin{enumerate}[leftmargin=*,labelsep=4.9mm]
    \item Signal \texttt{ini} is set to ``0'' and all registers are reset.
    \item First comparisons, $u_{2j-1}<u_{2j}$ with $j=1,\dotsc,M/2$, are computed and stored in each of the $M/2$ comparer registers.
    \item Signal \texttt{ini} is set back to ``1'' and the next comparison is performed. Consequently, comparisons are stored in registers 1 to $M/4$, while registers $M/4+1$ to $M/2$ are now filled with ones.
    \item Succesive $\left \lceil log_2 M \right \rceil-1$ comparisons are computed until a~valid result is obtained.
\end{enumerate}

\subsubsection{Internal~ROM}
The ROM, as marked in green in Figure~\ref{fig:arq_global}, stores the cluster to which each neuron belongs. The~cluster identification is performed by addressing the ROM while using the index of the BMU identified in the tree comparer module, consequently allowing one to know which cluster the input sample fits the most. The~ROM is implemented using look-up tables (LUTs) with the aim of improving the circuit speed. LUTs are typical FPGA resources that reduce the propagation delays when compared to block RAM-based~memories.

\subsubsection{Parameterization and Control~Signals}
The complete structure of the SOM HW accelerator is parametric and fully customizable. ROMs~containing the neurons' weights as well as the clusters that are associated with each neuron are simultaneously initialized. Elements such as type depths, signal bit-widths, and~the number of inputs and neurons were defined on a~standalone~package.

 The architecture's latency depends on two factors: a~fixed time that always delays the same number of clock cycles and~a variable time that depends on the number of features ($N$) and the number of neurons ($M$). Thus, one clock cycle is needed to load the input registers and two clock cycles for the computation of Euclidean distance within the neuron's distance~modules. 
 
 On the other hand, the~neurons' adder module computes its outputs 2 by 2.  Consequently,~the~clock cycles that are required to obtain the output of the adder module are calculated, as follows:
\begin{equation}
    t_{tree}=\left \lceil log_2 N \right \rceil = min \left\{ k \in \mathbb{Z} | \left( log_2 N \right)\leq k\right \}.
    \label{eq:comparer}
\end{equation}

As in the case of the adder module, the~tree comparer decides which neuron holds the shortest distance recursively two by two. For~that reason, its latency is computed in the same way as in Equation~\eqref{eq:comparer}, but~using $M$ instead of $N$, since it has the array of the outputs of the neurons as inputs. Therefore,~the~number of clock cycles elapsed since the input signal arrives in the architecture until a~valid output is provided is computed by the following expression:
\begin{equation}
    \label{eq:ncy}
    N_{cycles}=3+\left \lceil log_2 N\right \rceil + \left \lceil log_2 M \right \rceil.
\end{equation}

\begin{figure}[H]
    \centering
    \includegraphics[width=1\textwidth, height=0.48\textwidth]{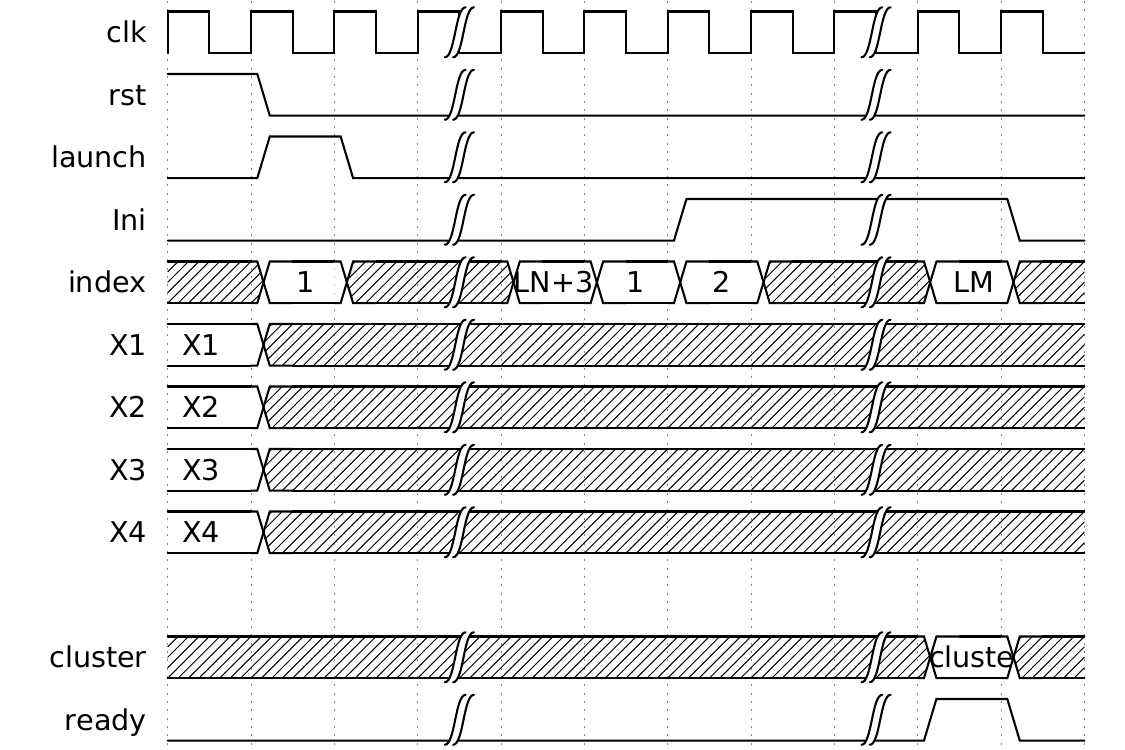}
    \caption{Chronogram of the control-signal sequence of the SOM HW accelerator. LN stands for $\left \lceil log_2 N\right \rceil$, and~LM for $\left \lceil log_2 M\right \rceil$.}
    \label{fig:control_applied}
\end{figure}

On the other hand, the~control signals of the SOM HW accelerator are \texttt{rst}, \texttt{launch}, and~\texttt{ini} (see~Figure~\ref{fig:control_applied}); they work, as follows:

\begin{enumerate}[leftmargin=*,labelsep=4.9mm]
    \item \texttt{rst} clears all of the architecture registers and prepares the modules for a~new input array.
    \item \texttt{launch} loads the input data into the neurons' input registers. Neurons' outputs are ready after $3+\left \lceil log_2 N\right \rceil$ clock cycles.
    \item In the next clock cycle, the~neurons' outputs are loaded into the input registers of the recursive tree comparer module.
    \item \texttt{Ini} is triggered to indicate to trigger the recursive comparisons of the tree adder. The~result is ready after $\left \lceil log_2 M \right \rceil$ clock cycles.
\end{enumerate}

\vspace{-2pt}

\subsection{FPGA~Implementation}

In Section~\ref{sec:SOM}, the~SOM training as well as a~classification method were explained. Additionally,~the~classification was carried out on the Uyanik dataset completed by the GT-Suite simulation data, obtaining the weights and clusters to which the neurons belong. With~that data, the~SOM network was implemented in the FPGA in order to classify the Uyanik drivers by storing the neuron weights and their corresponding clusters into the internal ROM of the~architecture.

The~FPGA of the Xilinx ZynQ-7000 family (XC7Z045-2FFG900 PSoC) \cite{DS190} is used in order to implement the SOM accelerator. This FPGA, derived from the Xilinx Kintex-7 family, has the following logic~resources:
\begin{itemize}[leftmargin=*,labelsep=5.8mm]
    \item  54 650 logic blocks, each one conformed by four six-input LUTs  and eight flip-flops;
    \item 19.2 Mbits of high-speed RAM blocks;
    \item 900 digital signal processing (DSP) blocks; and,
    \item an~analog-to-digital converter (ADC).
\end{itemize}

The architecture described in Section~\ref{subsec:HW} has a~total of $M=121$ neurons and $N=4$ input features. We used fixed-point binary arithmetic to represent data. Both input data and neurons' weights have been represented with 8 bits, with~all the 8 bits representing the decimal part, because~both are unsigned positive numbers. The~intermediate operations' bitwidths were selected such that neither overflows nor rollovers could~occur.

\subsubsection{Resource~Usage}

The full HW system was successfully implemented, with~the post-implementation results displayed in Table~\ref{tab:implementation}. The~SOM network acceleration system fit into the selected PSoC’s logic, leaving enough resources available for further system applications, scalations, or~improvements.

\begin{table}[H]
\caption{Post-implementation resources report (Xilinx XC7Z045-2FFG900).}
\label{tab:implementation}
\centering
\begin{tabular}{cccc}
\toprule
\textbf{Resource}	& \textbf{Utilization}	& \textbf{Available} & \textbf{\% Used}\\
\midrule
LUT	& \hl{21 107}	& \hl{218 600} & 9.66\\
Flip-flops	& \hl{13 337} & \hl{437 200} &3.05\\
\bottomrule
\end{tabular}
\end{table}
\unskip
%MDPI: Please confirm if it is necessary to add commas for numbers of more than four-digit/five in Table.
%Author: We added separations.

\subsubsection{Timing~Performance}
\label{sec:timing}
Before the implementation, the~maximum operational frequency was calculated. For~that purpose, the~architecture was implemented with a~minimum clock period of 10 ns, obtaining a~slack of 2.292 ns. Thus, the~maximum operational frequency of the design can be calculated, as follows:
\begin{equation}
    f_{max}=\frac{1}{T_{imp}-d_{slack}}=\frac{1}{10\text{ ns} - 2.292\text{ ns}}=129.74\text{ MHz}.
    \label{eq:fin_t}
\end{equation}

With the maximal clock frequency of Equation \eqref{eq:fin_t}, the~designed HW implementation could be used as an~AXI4 peripheral dependent on an~AXI4 bus clock frequency of 100 MHz. With~this operational frequency, and~applying Equation \eqref{eq:ncy} with $N=4$ features and $M=121$ neurons, this~design delayed 12 clock cycles (0.12 $\upmu$s at $\text{F}_{\text{CLK}}=100$ MHz) to return a~valid result with a~power consumption of 0.3 W. These results outperformed the timing that was obtained for a~full-software PC-based MATLAB model design (20-core Intel Xeon E5-2630 v4 CPU \@ 2.20 GHz with 32 GB of DDR4 RAM), with~a timing performance of 128.03~$\pm$~12.01 $\upmu$s to compute the same SOM, as~well as a~PC-based, C-coded prototype that achieved timing marks of 1.34 $\pm$ 0.28 $\upmu$s.

The obtained timing performance was better than in other FPGA-based SOM applications, such as the work by the authors of~\cite{Tisan2013}, where the highest operational frequency obtained was 101.54~MHz for a~considerably smaller SOM network. On~the other hand, in~\cite{kurdthongmee2016hardware}, an~absolutely novel architecture is presented with the aim of improving the general performance of the minimum's finding procedure by sidelining the use of comparers. This architecture, despite operating at even lower frequencies (a maximum of 19.6 MHz), achieves very high throughput at the cost of drastically increasing complexity. In~contrast, in~\cite{Hikawa2015}, outstanding frequencies of 188.9 MHz are achieved for a~bigger SOM implementation by carrying out some simplifications that might compromise the accuracy of smaller~networks.

Consequently, we can assert that the HW partition developed in this work is an~appropriate solution between conventional SW-based approaches and novel FPGA-based, extreme~performance architectures, which provides an~adequate trade-off between complexity, performance, and~development~time.

\subsubsection{Simulation~Results}

\begin{figure}[H]
    \centering
    \includegraphics[width=15.5 cm]{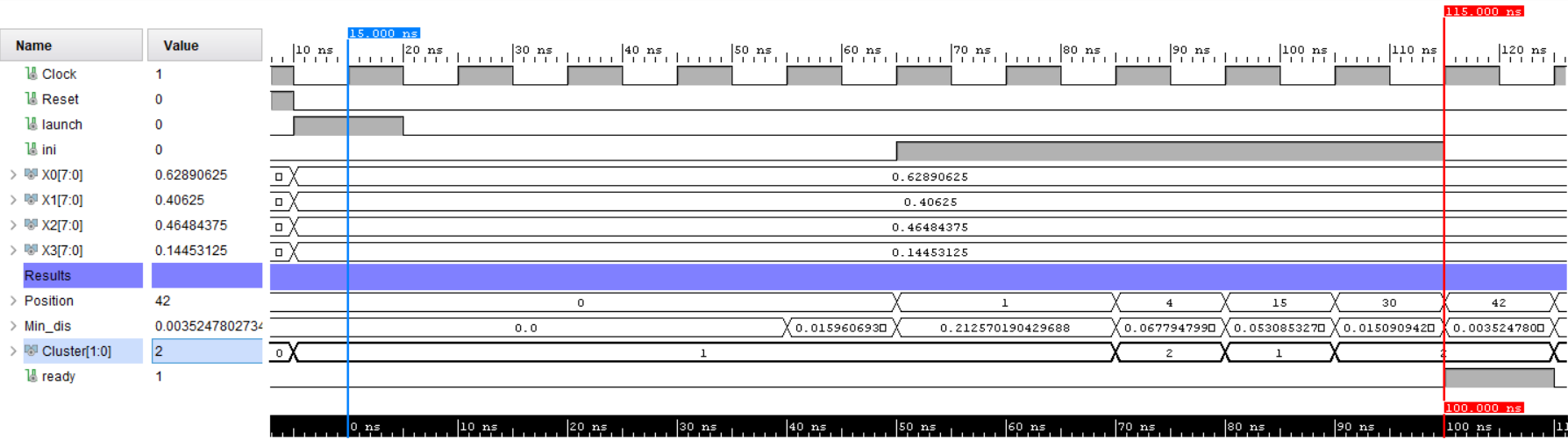}
    \caption{Simulation results of the SOM HW accelerator obtained with the Vivado design~suite.}
    \label{fig:simures}
\end{figure}

In order to verify that the developed SOM HW accelerator works as expected, several input arrays are fed to the architecture, the~control signals displayed in Figure~\ref{fig:control_applied} are applied to the circuit, and~the identified clusters are verified, as~can be seen in Figure~\ref{fig:simures}. 

In this figure, a~latency of 12 clock cycles (0.12 $\upmu$s) is shown, which mat with the timing indicated in Section~\ref{sec:timing}. Additionally, the~BMU's distance is checked and compared against an~equivalent MATLAB SOM model (with a~value of $3.52 \times 10^{-3}$), jointly with the cluster ROM position \mbox{(position = 42)}, which 
the HW, the fixed-point implementation, %please check
and the MATLAB floating-point model results totally~match.

\subsection{Software~Partition}

The processing system of the PSoC (SW partition) is built around a~dual-core ARM Cortex-A9 microprocessor. With~this architecture in mind, the~SW application was developed, enabling the full operation of the hybrid HW/SW system. These functionalities are as~follows:
\begin{itemize}[leftmargin=*,labelsep=5.8mm]
    \item I/O management: the system retrieves the driving features from the buses of the vehicle (e.g.,~CAN-bus)  and outputs the natural-language driver recommendations.
    \item Data logging and windowing: the microprocessor stores 8 s of data (256 samples \@ 32 Hz) to compute the data windows used to extract the driving features. Each window has an~overlapping of 4 s with its preceding one; thus,  with~an 8 s size, a~new window is generated every 4 s.
    \item Feature computation: for each data window, the~microprocessor computes the average values of ERPM, GP, and~PGP,  and~the variance of Pos XACC.
    \item Data exchange: the SW partition sends the features to the HW accelerator through the AXI4 bus and retrieves the identification results (DS clusters) from the HW partition.
    \item Cluster distribution computation: the cluster distribution of a~set of windows, evaluated during a~certain time of uninterrupted driving above a~certain speed threshold, is computed.
    \item Driver advice: natural-language advice is provided, depending on the cluster distribution, that is to say, according to the cluster in which the driver spends the longest~time.
    
\end{itemize}

In sum, with~the aforementioned data logging, windowing, and~feature computation, once the data are fed to the FPGA, the~SOM HW accelerator identifies the DS cluster for that window. With~those results, the~SW partition computes a~cluster distribution, decides which cluster registered the highest number of hits, and, according to that maximum, provides eco-driving advice to drivers.
%Consequently, the hybrid HW/SW implementation developed in this work is an~innovative solution between conventional SW-based approaches and novel FPGA-based, extreme performance architectures, which, provides an~adequate trade-off between complexity, performance, and development time.

\section{Concluding~Remarks}
\label{sec:concluding}

The main motivation of this work was the development of ADAS on the board vehicle contributing to the encouragement of eco-driving by providing real-time personalized advice to drivers. With~this aim, a~holistic approach, based on ML techniques and FPGA technology, was proposed. It uses a~data-based focus to identify relevant, fuel-consumption-associated features. A~mix of real-world data, which were obtained with an~instrumented car and~fuel consumption simulation results, was jointly processed. These~data were obtained from a~meaningful sample of motorists driving along the same route, under~similar conditions, to~assure the reproducibility of the results. Particularly,~stretches~registering the motorway or highway driving at least 60 km/h  were analyzed to faithfully represent high-speed driving with no outliers. This analysis had the goal of providing informative data to train an~SOM, which, after~a clustering process, allows for one to classify fuel-consumption-compromising DSs. The~clusters identified in this piece of research are representative of several fuel-consumption-related DSs. Hopefully, this cluster-based approach can be extrapolated to more complex driving contexts, such as driving in dense traffic or urban roads, due to its unsupervised~nature. 

The DS recommendations developed are designed to be valid for the majority of drivers. They~are provided using natural language, and can be easily understood and followed by most drivers. If~a~given driver follows the advice, he/she will increase in ecological awareness, modify his/her DS, and~consequently reduce fuel consumption and pollutant emissions, with~the expected results ranging from the 9.5\% to the 31.5\%, or~even better in the case of a~driver with a~high level of engagement with the advices that the system provides. In~addition, current implementations of efficient driving strategies for autonomous vehicles could also benefit from these results by incorporating the proposed system at the development stage, or~even after it, to~improve and verify the eco-friendliness of the developed~model.

The solution adopted in this work relies on a~high-performance, fully parallel SOM implementation.~This architecture is inherently parallelizable for high-performance hardware implementation due to its layered topology, and~easily scalable due to its extensive parametrization. The~entire SOM-based classification system was successfully implemented while using an~FPGA device of the Xilinx ZynQ-7000 PSoC family, with~the HW partition providing high speed and low-power consumption for real-time implementation, while its microprocessor executed complementary tasks. Moreover, due to the reconfigurable nature of FPGAs, both the hardware and software partition of the PSoC can be updated to cope with the continuous changes that new vehicle technologies introduce.%, including the challenges of performing these eco-advice at the rising of electric vehicles'~utilization.

In future work, SOM-based intelligent system applications will be broadened, adding more driving scenarios to those already researched for DS-related fuel consumption on highways and roads. It is worth remarking that further work can be done to decide which advice should be provided to drivers whose predominant DS is divided between non-contiguous clusters. Additionally,~we~plan to deploy this system in an~actual car, so as to test the engagement of real drivers with the recommendations provided by the system, as~well as their effects on actual fuel consumption reduction compared to built-in ECO modes. Finally, estimations of NOx, CO, and~HC emissions will be performed to further analyze the effects of the system presented in this~work.

\vspace{6pt} 

%%%%%%%%%%%%%%%%%%%%%%%%%%%%%%%%%%%%%%%%%%
%% optional
%\supplementary{The following are available online at \linksupplementary{s1}, Figure S1: title, Table S1: title, Video S1: title.}

% If you wish to submit a~video article, please do so with any other supplementary material.

%%%%%%%%%%%%%%%%%%%%%%%%%%%%%%%%%%%%%%%%%%
\authorcontributions{Conceptualization, Ó.M.-C, M.D.-R, I.d.C. and V.M.; methodology,  Ó.M.-C, M.D.-R, I.d.C. and V.M.; software,  Ó.M.-C, M.D.-R, I.d.C. and V.M.; validation,  Ó.M.-C, M.D.-R, I.d.C. and V.M.; investigation, Ó.M.-C, M.D.-R, I.d.C. and V.M.; writing—original draft preparation, Ó.M.-C. and I.d.C.; writing—review and editing, Ó.M.-C. and I.d.C. All authors have read and agreed to the published version of the manuscript.}
%MDPI: Please confirm to add more specific information.
%For research articles with several authors, a short paragraph specifying their individual contributions must be provided. The following statements should be used ``Conceptualization, X.X. and Y.Y.; methodology, X.X.; software, X.X.; validation, X.X., Y.Y. and Z.Z.; formal analysis, X.X.; investigation, X.X.; resources, X.X.; data curation, X.X.; writing--original draft preparation, X.X.; writing--review and editing, X.X.; visualization, X.X.; supervision, X.X.; project administration, X.X.; funding acquisition, Y.Y. All authors have read and agreed to the published version of the manuscript.'', please turn to the  \href{http://img.mdpi.org/data/contributor-role-instruction.pdf}{CRediT taxonomy} for the term explanation. Authorship must be limited to those who have contributed substantially to the work reported.

%%%%%%%%%%%%%%%%%%%%%%%%%%%%%%%%%%%%%%%%%%
\funding{This work was supported in part by the Spanish AEI and European FEDER funds under Grant TEC2016-77618-R (AEI/FEDER, UE) and by the University of the Basque Country under Grant GIU18/122.}

%%%%%%%%%%%%%%%%%%%%%%%%%%%%%%%%%%%%%%%%%%
%\acknowledgments{In this section you can acknowledge any support given which is not covered by the author contribution or funding sections. This may include administrative and technical support, or donations in kind (e.g., materials used for experiments).}

%%%%%%%%%%%%%%%%%%%%%%%%%%%%%%%%%%%%%%%%%%
\conflictsofinterest{The authors declare that there is no conflict of~interest.} 

\reftitle{References}
%\bibliography{biblio}

% The following MDPI journals use author-date citation: Arts, Econometrics, Economies, Genealogy, Humanities, IJFS, JRFM, Laws, Religions, Risks, Social Sciences. For those journals, please follow the formatting guidelines on http://www.mdpi.com/authors/references
% To cite two works by the same author: \citeauthor{ref-journal-1a} (\citeyear{ref-journal-1a}, \citeyear{ref-journal-1b}). This produces: Whittaker (1967, 1975)
% To cite two works by the same author with specific pages: \citeauthor{ref-journal-3a} (\citeyear{ref-journal-3a}, p. 328; \citeyear{ref-journal-3b}, p.475). This produces: Wong (1999, p. 328; 2000, p. 475)

%=====================================
% References, variant B: external bibliography
%=====================================
%\externalbibliography{yes}
%\bibliography{biblio.bib}

%%%%%%%%%%%%%%%%%%%%%%%%%%%%%%%%%%%%%%%%%%
%% optional
%\sampleavailability{Samples of the compounds ...... are available from the authors.}

%% for journal Sci
%\reviewreports{\\
%Reviewer 1 comments and authors’ response\\
%Reviewer 2 comments and authors’ response\\
%Reviewer 3 comments and authors’ response
%}

%%%%%%%%%%%%%%%%%%%%%%%%%%%%%%%%%%%%%%%%%%
\end{document}